\UseRawInputEncoding
\documentclass[journal]{IEEEtran}
\usepackage{graphicx}
\usepackage{subfigure}
\usepackage{booktabs}
\usepackage{cite}
\usepackage{caption}

\usepackage{float}
\usepackage{multirow}
\usepackage{color}
\usepackage{algorithm}
\usepackage{algorithmic}
\usepackage{amsmath}

\begin{document}
    \title{Deep Lifelong Cross-modal Hashing}
    \author{Liming Xu, Hanqi Li, Bochuan Zheng, Weisheng Li, Jiancheng Lv
    \thanks{Liming Xu is the corresponding author. Liming Xu, Hanqi Li and Bochuan Zheng are with the School of Computer Science, China West Normal University, Nanchong 637009, China. Liming Xu and Jiancheng Lv are with the College of Computer Science, Sichuan University, Chengdu 610065, China. Weisheng Li is with the College of Computer Science and Technology, Chongqing University of Posts and Telecommunication, Chongqing 400065, China. E-mail: xulimmail@gmail.com; lihq@gmail.com; zhengbc@cwnu.edu.cn; liws@cqupt.edu.cn; lvjiancheng@scu.edu.cn.}}
    \markboth{}
{Shell \MakeLowercase{\textit{Liming Xu et al.}}: Bare Demo of IEEEtran.cls for IEEE Journals}

\maketitle

\begin{abstract}
    Hashing methods have made significant progress in cross-modal retrieval tasks with fast query speed and low storage cost. Among them, deep learning-based hashing achieves better performance on large-scale data due to its excellent extraction and representation ability for nonlinear heterogeneous features. However, there are still two main challenges in catastrophic forgetting when data with new categories arrive continuously, and time-consuming for non-continuous hashing retrieval to retrain for updating. To this end, we, in this paper, propose a novel deep lifelong cross-modal hashing to achieve lifelong hashing retrieval instead of re-training hash function repeatedly when new data arrive. Specifically, we design lifelong learning strategy to update hash functions by directly training the incremental data instead of retraining new hash functions using all the accumulated data, which significantly reduce training time. Then, we propose lifelong hashing loss to enable original hash codes participate in lifelong learning but remain invariant, and further preserve the similarity and dis-similarity among original and incremental hash codes to maintain performance. Additionally, considering distribution heterogeneity when new data arriving continuously, we introduce multi-label semantic similarity to supervise hash learning, and it has been proven that the similarity improves performance with detailed analysis. Experimental results on benchmark datasets show that the proposed methods achieves comparative performance comparing with recent state-of-the-art cross-modal hashing methods, and it yields substantial average increments over 20\% in retrieval accuracy and almost reduces over 80\% training time when new data arrives continuously.
\end{abstract}

\begin{IEEEkeywords}
    Lifelong hashing, cross-modal retrieval, distribution heterogeneity, catastrophic forgetting, multi-label semantic similarity.
\end{IEEEkeywords}

\section{Introduction}
    \IEEEPARstart{D}{aily} multimedia data associated with potential semantic correlation shows multi-source heterogeneous and multi-modal characteristics. Based on the correlation, cross-modal retrieval methods\cite{1} aim to achieve cross-matching between different modalities. Cross-modal hashing\cite{2}, an important branch in cross-modal retrieval, maps high-dimensional original data into low-dimensional compact binary hash codes in Hamming space while maintaining similarity in original space by enforcing two instances with smaller (larger) Hamming distance be more (less) similar. Then, many cross-modal hashing methods have been proposed and it can generally divided into hand-craft and deep learning\cite{3} based methods. Due to the data-driven characteristics and excellent feature extraction ability, we mainly focus on the deep learning based cross-modal hashing methods.

    Although existing deep cross-modal hashing methods have achieved satisfactory performance, there are still two main challenges: (1) catastrophic forgetting when incremental data with new categories is added and (2) non-continuity of hashing retrieval. Specifically, when incremental data with new category arriving continuously, it will yield catastrophic forgetting\cite{4}, i.e., the performance on learned hashing function significantly degrades over time as incremental data with new categories is added. From subsequent results, it can be estimated that when adding two categories data, the compared deep methods incur over 20\% accuracy drop. Then, almost all deep cross-modal hashing need to retrain hash functions using all the data to generate new hash codes when new data arrives, which can be time-consuming and inefficient. The subsequent results show that these methods require more training time to retrain the accumulated data, and it will be even worse on larger dataset. Besides, the consideration of distribution heterogeneity when new data arriving is insufficient. Most current cross-modal hashing use single-label which considers two instances as similar if they share at least one label to construct semantic similarity, which cannot preserve semantic similarity in original space and Hamming space effectively.

    Taking the above issues into consideration, we propose a novel \underline{D}eep \underline{L}ifelong \underline{C}ross-modal \underline{H}ashing method(DLCH) by incorporating the base learning and lifelong learning. Within it, lifelong hashing learning is design to directly learn and update hashing function of incremental data with new categories. Then, multi-label semantic similarity and self-supervised strategy are proposed to obtain high-quality hash codes. The main contributions of this paper can be summarized as follows:
        \begin{itemize}
                \item We propose a novel deep lifelong cross-modal hashing method to overcome catastrophic forgetting when incremental data with new categories is added. The original hash codes keep unchanged during incremental hash codes learning, the model performance on original data will not degrade after training on the incremental data.
                \item We design lifelong hashing loss to directly learn incremental hash codes instead of retraining the learned hashing functions, and optimize incremental hash codes bit by bit, which avoids huge time cost caused by retraining all the accumulated data.
                \item We define multi-label semantic similarity with multi-label information to describe semantic correlation more accurately and generate high-quality original hash codes. The detailed analysis is provided to show that it improves performance especially in the lifelong hashing.
                \item Extensive experiments on three benchmark cross-modal datasets demonstrate that the proposed DLCH gains competitive retrieval performance and obtains significant time reduction comparing with the state-of-the-art methods.
        \end{itemize}

    The rest of this paper is organized as follows: Section II reviews related work on cross-modal hashing retrieval. Section III describes the proposed deep lifelong cross-modal hashing method and optimization. Section IV presents the experimental results and analysis. Finally, Section V concludes our work.

\section{Related Work}
    Generally, the existing cross-modal hashing methods can be roughly categorized into hand-crafted and deep learning-based methods according to features extraction\cite{5, 6}. The former learns hash functions by hand-crafted features and shallow architecture, while the latter embeds deep network and fine-tunes training in an end-to-end way. We mainly focus on the deep learning-based cross-modal hashing, and summarizes the related issues in the view of continuity as follows.

    \subsection{Non-continuous Cross-modal Hashing}
        In the view of continuity, most of the existing cross-modal hashing methods are non-continuous, which means all training data need be provided at the beginning and loaded at once, and it will be re-trained when incremental data is added. As reported in previous works\cite{1, 2}, the deep learning-based cross-modal hashing mainly contains deep neural networks(DNN)- and Generative adversarial networks(GAN)-based cross-modal hashing according to different deep models.

        Furthermore, according to whether supervised information is involved, DNN-based methods can be divided into the unsupervised and the supervised. Specifically, unsupervised learns hash codes by exploring the correlation of heterogeneous data with correlation graphs or latent semantic space. DCSH\cite{7} adopts spectral clustering and anchor-to-anchor mapping for better similarity graphs. To bridge modality gap, JDSH\cite{8} exploits a joint-modal similarity matrix, while DUCMH\cite{9} relies on data alignment and image-text data pairs. DGCPN\cite{10} explores intrinsic semantic relationships with graph-neighbor coherence to avoid suboptimal retrieval Hamming space. With introducing knowledge distillation scheme, KDCMH\cite{11} trains an unsupervised method as the teacher model used to provide distillation information to guide supervised method. CMIMH\cite{12} tries to find a balance between reducing the modality gap and losing modality-private information by maximizing mutual information. UCHM\cite{13} focuses on image-text interaction to generate a superior modality-interaction-enabled similarity matrix for training set. On the other hand, supervised methods make full use of semantic information in supervised information(e.g. tags or pair-wise similarity information) and generally achieve better accuracy. To highlight useful information and suppress redundant information, TEACH\cite{14} and MMACH\cite{15} add attention mechanism to feature learning process, and the latter utilizes multi-label information to further improve accuracy additionally. HSSAH\cite{16} replaces binary similarity matrix by asymmetric high-level semantic similarity to maintain richer semantic information. For data pairs with different similarities, DCH-SCR\cite{17} fine-tunes weights of positive instances to vary optimization strengths.

        In addition to the DNN-based methods described above, GANs\cite{18} are also often used for cross-modal hashing  retrieval. MLCAH\cite{19} proposes a global and local semantic alignment mechanism to encode multi-level correlation information into hash codes. DADH\cite{20} employs a weighted cosine triplet constraint to learn the ranking based similarity relevance of data points. MGAH\cite{21} fits the underlying manifold structure using a multi-pathway generative adversarial network. CPAH\cite{22} utilizes consistency refined module to implement the separation of incompatible modality-common and modality-private representations. DAGNN\cite{23} extracts more representative common feature vectors with the aid of the dual generative adversarial networks and the multi-hop graph neural networks. SAAH\cite{24} depends on inter-modal and intra-modal adversarial autoencoder modules to generate uniform feature representation. DFAH\cite{25} overcomes semantic gap and distribution shift via adversarial learning between Modality Discriminator and Modality-Specific Feature Extractor. CMGCAH\cite{26} introduces a transformer-based feature extraction network for further leveraging position information.

        Despite the much progress made by the mentioned deep cross-modal hashing methods, the whole model and hash function will be re-trained once data changes due to its non-continuity and the nature of catastrophic forgetting, which causes much training time and resource cost.

    \subsection{Continuous Cross-modal Hashing}
        Considering that the above non-continuous models are not suitable for large-scale data sets due to large resource cost when training streaming data, some have embedded online learning processing into hash learning. To this end, OCMFH\cite{27} uses collective matrix factorization to learn hash codes for streaming data and updates hash codes of old data according to dynamic changes of hash model without accessing to old data. Similarly, OCMH\cite{28} transforms inefficient update of hash codes into efficient update of shared latent encoding matrix and dynamic transition matrix. By taking the label information into account, OLSH\cite{29} maps discrete labels to continuous latent semantic concept space, on which similarity measures between data points are performed. Then, LEMON\cite{30} designs label embedding framework to generate discriminative hash codes, which makes full use of semantic information. OLCH\cite{31} introduces online semantic representation learning strategy to preserve the similarity between new data and old data in Hamming space. Besides, in order to avoid quantization error, DOCH\cite{32} optimizes discretely binary constraints and yields uniform high-quality hash codes. OMGH\cite{33} utilizes anchor-based manifold embedding to sparsely represent old data and adaptively guide hash learning.

        Although these hashing methods attempts to introduce the online learning process to improve computational efficient and reduce resource cost, they focus on optimization and do not eliminate the limitation that the old hash codes will be changed. Then, it can be clearly found that the existing online cross-modal hashing methods are shallow architectures, and there is current no deep lifelong cross-modal hashing method. It's well known that the deep neural network is capable to capture the non-linear heterogeneous features. Thirdly, these online cross-modal hashing has considered to save computational cost when incremental data is added continuously, but they ignore the distribution change which will cause server catastrophic forgetting and fail to reach high performance when new categories appear. Additionally, the retrieved performance can be limited by using the simple single-label similarity evaluation.

        Motivated by the weaknesses of existing methods, we propose deep lifelong cross-modal hashing to avoid catastrophic forgetting and reduce training time. Firstly, we design lifelong learning strategy to directly learn hash codes of incremental data on the basis of unchanging old hash codes instead of retraining hash functions using all the accumulated data. Then, we incorporate lifelong hashing loss preserve the similarity and dis-similarity among the original and incremental hash codes, which is excepted to overcome catastrophic forgetting and performance drops. Thirdly, we introduce multi-label semantic similarity to supervise hash learning to further utilize label information to adapt distribution heterogeneity and improve accuracy. These three composition constitute the proposed DLCH.

\section{The proposed method}
    In this section, the details of DLCH are presented. Fig.1 illustrates the framework overview, which includes image and text modality for convenient description. We have to note that our proposed method can easily be extended to more modalities (e.g., image, text, audio and graphics).
        \begin{figure*}[!t]
            \centering
            \includegraphics[width=1\linewidth]{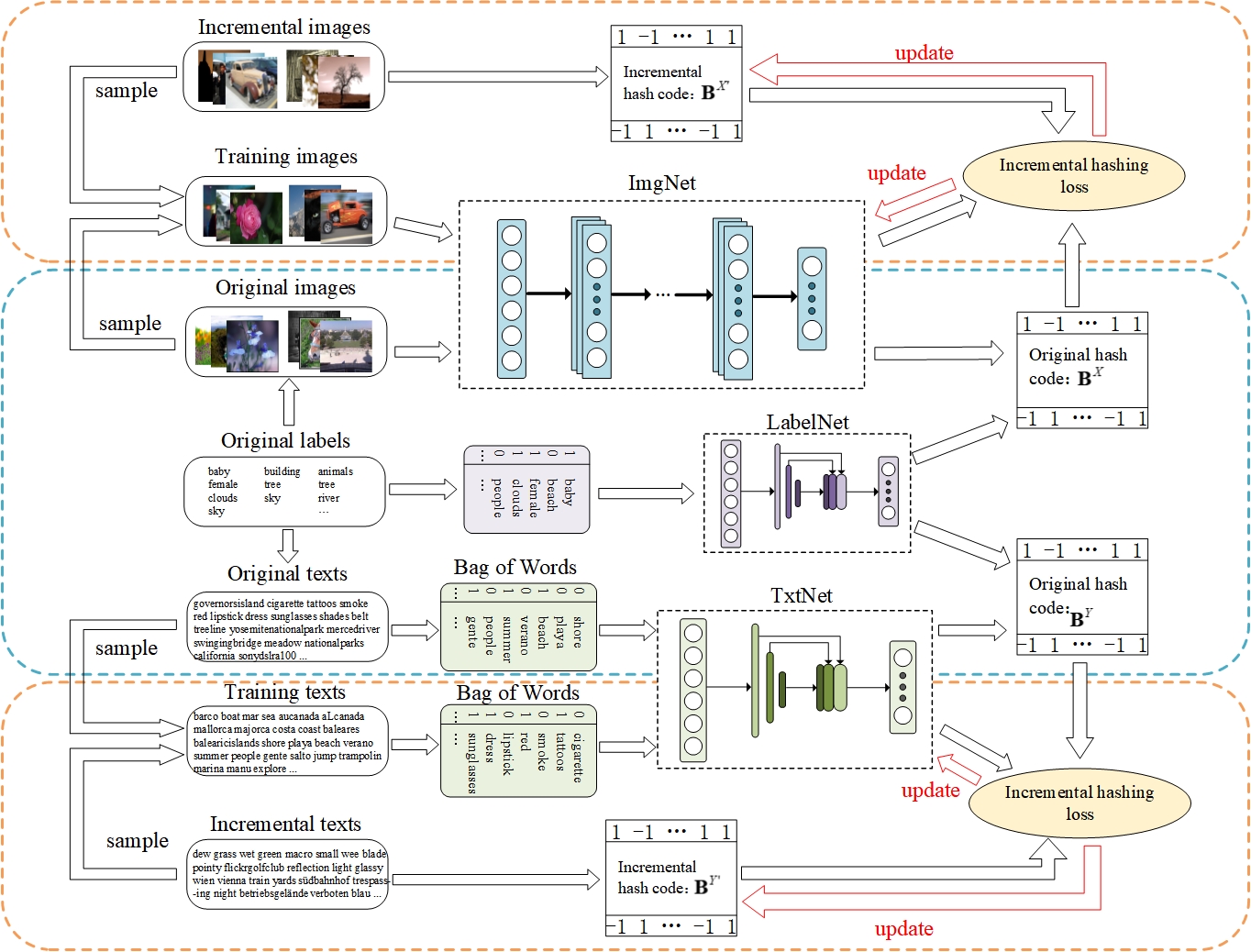}
            \caption{\small{DLCH consists of original and lifelong hash learning phase. Original hash learning phase (blue dashed box) where LabelNet acts as supervisor to guide hash function learning learns hash function as previous works conduct. Lifelong learning phase (orange dashed box) encompasses three steps: (1) sample training data; (2) construct lifelong hashing loss, and (3) update incremental hash codes.}}
            \label{Fig1}
        \end{figure*}

    \subsection{Notations and Problem Definition}
        It is necessary to present notations and problem definition firstly. In this paper, bold uppercase letters, like ${\bf{W}}$, denote matrices; italic lowercase letters, like ${w}$, denote vectors. ${{\bf{W}}_{i*}}$ denotes the ${i}$-th row of matrix ${\bf{W}}$, ${{\bf{W}}_{*j}}$ denotes the ${j}$-th column of matrix ${\bf{W}}$, and ${{\bf{W}}_{ij}}$ denotes the element of the ${i}$-th row and ${j}$-th column of ${\bf{W}}$. The transpose of ${\bf{W}}$ is denoted as ${{\bf{W}}^{\rm{T}}}$ and $tr({\bf{W}})$ is the trace of ${\bf{W}}$. $\left\| {\bf{W}} \right\|_F^2 = tr({{\bf{W}}^{\rm{T}}}{\bf{W}})$ is the Frobenius-norm. $sign\left( {\rm{\cdot}} \right)$ is sign function defined as:
            \begin{equation}\label{eq1}
                sign(x) = \left\{ \begin{array}{c}
                1,\quad\quad x \ge 0\\
                 - 1,\quad x < 0
                \end{array} \right.
            \end{equation}

        Suppose there are ${m}$ labeled instances ${\bf{O}} = \{ {\bf{X,Y,L}}\} $ in the database, where ${\bf{X}}=\{{x_i}\}_{i=1}^m\in{{\bf{R}}^{m\times{d_x}}}$, ${\bf{Y}}=\{{y_i}\}_{i=1}^m\in{{\bf{R}}^{m\times{d_y}}}$ and ${\bf{L}}=\{{l_i}\}_{i=1}^m\in{\{0,1\}^{m\times c}}$ denote image, text and label, respectively. ${d_x}$ and ${d_y}$ denote the dimension of original image and text features, respectively. \emph{c} is the total number of original classes. That is, an instance ${o_i}$ associated with class label ${l_i}$ has both image feature ${x_i}$ and text feature ${y_i}$. If ${x_i}$ or ${y_i}$  belongs to the ${j}$-th class, then ${l_{ij}} = 1$, otherwise, ${l_{ij}} = 0$. 

        For these ${m}$ instances, our goal is to learn multi-label network $h({l_i}{\rm{;}}{\theta _l})$, image hash function $f({x_i}{\rm{;}}{\theta _x})$ and text hash function $g({y_i}{\rm{;}}{\theta _y})$ to generate the corresponding hash representation ${\bf{H}} = \{ {{\bf{H}}_i}|{{\bf{H}}_i} = h({l_i};{\theta _l})\}  \in {{\bf{R}}^{k \times m}}$, ${\bf{F}} = \{ {{\bf{F}}_i}|{{\bf{F}}_i} = f({x_i};{\theta _x})\}  \in {{\bf{R}}^{k \times m}}$ and ${\bf{G}} = \{ {{\bf{G}}_i}|{{\bf{G}}_i} = g({y_i};{\theta _y})\}  \in {{\bf{R}}^{k \times m}}$, where \emph{k} denotes the length of hash codes. Subsequently, we can utilize above hash representation to obtain final original hash codes ${{\bf{B}}^X} \in {\{  - 1, + 1\} ^{k \times m}}$ and ${{\bf{B}}^Y} \in {\{  - 1, + 1\} ^{k \times m}}$.

        Analogously, when ${n}$ incremental labeled data ${\bf{O}}' = \{ {\bf{X}}'{\bf{,Y}}'{\bf{,L}}'\} $ arriving, each instance has also image modality ${\bf{X}}' = \{ {x_i}\} _{i = m + 1}^{m + n}$, text modality ${\bf{Y}}' = \{ {y_i}\} _{i = m + 1}^{m + n}$ and label ${\bf{L}}' = \{ {l_i}\} _{i = m + 1}^{m + n} \in {\{ 0,1\} ^{n \times c'}}$, where $c'$ is the number of new classes. For these new ${n}$ instances, we expect to update hash function $f({x_i}{\rm{;}}{\theta _x})$ and $g({y_i}{\rm{;}}{\theta _y})$, and gain corresponding binary hash codes ${{\bf{B}}^{X'}} \in {\{  - 1, + 1\} ^{k \times n}}$ and ${{\bf{B}}^{Y'}} \in {\{  - 1, + 1\} ^{k \times n}}$ with original hash codes invariant.

        To reach the goal, we firstly sample dataset ${a}$ from original dataset ${\bf{O}}$ and incremental dataset ${\bf{O}}'$ and name as training set ${{\bf{O}}_a} = \{ {{\bf{X}}_a}{\bf{,}}{{\bf{Y}}_a}{\bf{,}}{{\bf{L}}_a}\} $, where ${{\bf{X}}_a} \in {{\bf{R}}^{a \times {d_x}}} \subset \{ {\bf{X}}{\rm{,}}{\bf{X}}'\} $, ${{\bf{Y}}_a} \in {{\bf{R}}^{a \times {d_y}}} \subset \{ {\bf{Y}}{\rm{,}}{\bf{Y}}'\} $, ${L_a} = \{ {l_i}\} _{i = 1}^a \in {\{ 0,1\} ^{a \times (c + c')}}$, and $a = {a_1} + {a_2}$. ${a_1}$ and ${a_2}$ denote the number of training data sampled from original and incremental dataset, respectively. The main notations in this paper are summarized in Table II.
            \begin{table}[!t]
            \caption{\small{Notations and the Meanings.}}
            \label{tab2}
            \centering
            \begin{tabular}{|c|c|}
            \hline
                \multicolumn{1}{|c|}{Notation} & \multicolumn{1}{c|}{Meaning}                   \\  \hline\hline
                $\bf{X}$ ($\bf{Y}$)     & The image (text) features for original data.   \\ \hline
                $\bf{X'}$ ($\bf{Y'}$)   & The image (text) features for incremental data.\\ \hline
                $\bf{L}$ ($\bf{L'}$)    & The labels for original (incremental) data.\\
                \hline
        		$\bf{F}$/($\bf{G}$/$\bf{H}$)   & The hash representation of image/text/label modality.\\ \hline
        		${{\bf{B}}^X}$(${{\bf{B}}^Y}$)   & The original hash codes for images (texts).\\ \hline
        		${{\bf{B}}^{X'}}$(${{\bf{B}}^{Y'}}$)   & The incremental hash codes for images (texts).\\ \hline
                ${{\bf{A}}^X}$(${{\bf{A}}^Y}$)   & The image (text) hash codes of training data.\\\hline
                ${\bf{S}}$  & The multi-label semantic similarity matrix.\\ \hline
                ${\bf{Q}}$  & The hash representation similarity matrix.\\ \hline
        		${m}$   & Number of existing instances with original classes.\\ \hline
                ${n}$   & Number of instances with new classes.\\  \hline
                ${a}$   & Number of training instances.\\  \hline
        		${c}$ (${c}'$)     & Number of original (incremental) classes.\\  \hline
        		${k}$     & Number of hash codes bits.\\  \hline
            \end{tabular}
            \end{table}

    \subsection{Original Hash Codes Learning}
          \subsubsection{Original Hashing Loss}
            In original learning phase, our goal is to train LabelNet to supervise the training and to train ImgNet and TxtNet to output hash representation which preserves the similarity in original space. Thus, for given hash representation \textbf{F} of ImgNet and hash representation \textbf{H} of LabelNet, hash representation similarity is defined as:
                \begin{equation}\label{eq2}
                  {\bf{\bar Q}}_{ij}^{xl} = \frac{{{{\bf{F}}_i}}}{{{{\left\| {{{\bf{F}}_i}} \right\|}_2}}} \cdot {(\frac{{{{\bf{H}}_j}}}{{{{\left\| {{{\bf{H}}_j}} \right\|}_2}}})^{\rm{T}}}
                \end{equation}

            In order to unify the range of multi-label semantic similarity and hash representation similarity, ReLu transformation is applied to ${\bf{\bar Q}}_{ij}^{xl}$ which belongs to $[ - 1,1]$. Then, the hash representation similarity can be rewritten as:
                \begin{equation}\label{eq3}
                  {\bf{Q}}_{ij}^{xl} = max(0,{\bf{\bar Q}}_{ij}^{xl})
                \end{equation}

            Then we define inter-modality similarity preserving loss which measures the gap between multi-label semantic similarity and hash representation similarity of two instances from different modalities as:
            \begin{equation}\label{eq4}
                \begin{array}{c}
                    {J_{inter}} = J_{inter}^{xl} + \alpha J_{inter}^{yl}\\
                     = \left\| {{{\bf{S}}^{xl}} - {{\bf{Q}}^{xl}}} \right\|_F^2 + \alpha \left\| {{{\bf{S}}^{yl}} - {{\bf{Q}}^{yl}}} \right\|_F^2
                    \end{array}
            \end{equation}
            where $J_{inter}^{xl}$ ($J_{inter}^{yl}$) is the similarity preserving loss for image (text) modality and multi-label. ${\bf{S}}^{xl}$ and ${\bf{Q}}^{xl}$ are multi-label semantic similarity and hash representation similarity for image feature $\emph{x}$ and label $\emph{l}$, respectively. ${\bf{S}}^{yl}$ and ${\bf{Q}}^{yl}$ are multi-label semantic similarity and hash representation similarity for image feature $\emph{y}$ and label $\emph{l}$, respectively. The optimal hash representation $\bf{F}$ and $\bf{G}$ can be obtained by minimizing Eq.(4).

            Correspondingly, to measure the gap between multi-label semantic similarity and hash representation similarity of two instances from the same modality, we define intra-modality similarity preserving loss as:
            \begin{equation}\label{eq5}
              \begin{array}{c}
                {J_{intra}} = J_{intra}^{ll} + J_{intra}^{xx} + J_{intra}^{yy}\\
                 = \left\| {{{\bf{S}}^{ll}} - {{\bf{Q}}^{ll}}} \right\|_F^2 + \left\| {{{\bf{S}}^{xx}} - {{\bf{Q}}^{xx}}} \right\|_F^2 + \left\| {{{\bf{S}}^{yy}} - {{\bf{Q}}^{yy}}} \right\|_F^2
                \end{array}
            \end{equation}
            where $J_{intra}^{ll}$, $J_{intra}^{xx}$ and $J_{intra}^{yy}$ are the intra-modality similarity preserving loss for multi-label, image and text modality, respectively. ${\bf{S}}^{ll}$, ${\bf{S}}^{xx}$ and ${\bf{S}}^{yy}$ are multi-label semantic similarity for these three modalities. ${\bf{Q}}^{ll}$, ${\bf{Q}}^{xx}$ and ${\bf{Q}}^{yy}$ are corresponding hash representation similarity. By minimizing Eq.(5), the original similarity between instances from same modality can be preserved in binary Hamming space.

            In order to enable the output of hash function tends to -1 or +1 to obtain distinctive binary hash codes, we further add quantization loss which gaps difference between discrete binary hash codes and continuous hash representation as:
                \begin{equation}\label{eq6}
                      \begin{array}{c}
                        {J_{quan}} = \left\| {{\bf{F}} - {{\bf{B}}^X}} \right\|_F^2 + \left\| {{\bf{G}} - {{\bf{B}}^Y}} \right\|_F^2\\
                                   + \frac{1}{2}(\left\| {{\bf{H}} - {{\bf{B}}^X}} \right\|_F^2 + \left\| {{\bf{H}} - {{\bf{B}}^Y}} \right\|_F^2)
                      \end{array}
                \end{equation}

            Subsequently, we adopt bit-by-bit optimization to yield optimal hash codes following the previous works. Combining Eq.(4)-(6), we obtain the following final objective loss function:
                \begin{equation}\label{eq7}
                  \begin{array}{c}
                    \mathop {\min J}\limits_{{{\bf{B}}^X},{{\bf{B}}^Y},{\theta _x},{\theta _y},{\theta _l}}  = {J_{inter}} + \beta {J_{intra}} + \gamma {J_{quan}}\\
                    s.t. {\rm{   }}{{\bf{B}}^X} \in {\{  - 1, + 1\} ^{k \times m}}\\
                    {{\bf{B}}^Y} \in {\{  - 1, + 1\} ^{k \times m}},
                 \end{array}
                \end{equation}
            where $\beta$ and $\gamma$ are hyper-parameters which control the weight ratio of losses.

          \subsubsection{Optimization}
              We optimize parameters ${\theta _x}$, ${\theta _y}$ and ${\theta _l}$, and learn original hash codes ${{\bf{B}}^X}$ and ${{\bf{B}}^Y}$ with an alternating strategy. The detailed steps are provided below.

              \textbf{Step 1:} Learn ${\theta _l}$, with ${{\bf{B}}^X}$, ${{\bf{B}}^Y}$, ${\theta _x}$ and ${\theta _y}$ fixed

                When training LabelNet, we rewrite Eq.(7) as follows:
                  \begin{equation}\label{eq8}
                    \begin{array}{c}
                        \mathop {min}\limits_{{\theta _l}} J = \left\| {{{\bf{S}}^{xl}} - {{\bf{Q}}^{xl}}} \right\|_F^2 + \alpha \left\| {{{\bf{S}}^{yl}} - {{\bf{Q}}^{yl}}} \right\|_F^2 + \beta \left\| {{{\bf{S}}^{ll}} - {{\bf{Q}}^{ll}}} \right\|_F^2\\
                         + \frac{\gamma }{2}(\left\| {{\bf{H}} - {{\bf{B}}^X}} \right\|_F^2 + \left\| {{\bf{H}} - {{\bf{B}}^Y}} \right\|_F^2)
                        \end{array}
                  \end{equation}

                With ${{\bf{B}}^X}$, ${{\bf{B}}^Y}$, ${\theta _x}$ and ${\theta _y}$ fixed, we use stochastic gradient descent (SGD) and back-propagation (BP) algorithm to learn the parameter of multi-label network. The parameter gradient is calculated by
                    \begin{equation}\label{eq9}
                        \begin{array}{c}
                        \frac{{\partial J}}{{\partial {{\bf{H}}_{*j}}}} = 2({{\bf{Q}}^{xl}} - {{\bf{S}}^{xl}})\frac{{\partial {\bf{Q}}_{ij}^{xl}}}{{\partial {{\bf{H}}_{*j}}}} + 2\alpha ({{\bf{Q}}^{yl}} - {{\bf{S}}^{yl}})\frac{{\partial {\bf{Q}}_{ij}^{yl}}}{{\partial {{\bf{H}}_{*j}}}}\\
                         + 2\beta ({{\bf{Q}}^{ll}} - {{\bf{S}}^{ll}})\frac{{\partial {\bf{Q}}_{ij}^{ll}}}{{\partial {{\bf{H}}_{*j}}}} + \gamma ({{\bf{H}}_{*j}} - {\bf{B}}_{*j}^X) + \gamma ({{\bf{H}}_{*j}} - {\bf{B}}_{*j}^Y)
                        \end{array}
                    \end{equation}
                where
                    \begin{equation}\label{eq10}
                    \frac{{\partial {\bf{Q}}_{ij}^{xl}}}{{\partial {{\bf{H}}_{*j}}}} = \left\{ \begin{array}{c}
                    \frac{{{{\bf{F}}_i}}}{{{{\left\| {{{\bf{F}}_i}} \right\|}_2} \cdot {{\left\| {{{\bf{H}}_j}} \right\|}_2}}}(\frac{{\partial {\bf{H}}_j^{\rm{T}}}}{{\partial {{\bf{H}}_{*j}}}}),if\,{\bf{\bar Q}}_{ij}^{xl} \ge 0\\
                    0\quad \quad \quad \quad \quad \quad \quad ,if\,{\bf{\bar Q}}_{ij}^{xl} < 0
                    \end{array} \right.
                    \end{equation}
                    \begin{equation}\label{eq11}
                    \frac{{\partial {\bf{Q}}_{ij}^{yl}}}{{\partial {{\bf{H}}_{*j}}}} = \left\{ \begin{array}{c}
                    \frac{{{{\bf{G}}_i}}}{{{{\left\| {{{\bf{G}}_i}} \right\|}_2} \cdot {{\left\| {{{\bf{H}}_j}} \right\|}_2}}}(\frac{{\partial {\bf{H}}_j^{\rm{T}}}}{{\partial {{\bf{H}}_{*j}}}}),if\,{\bf{\bar Q}}_{ij}^{yl} \ge 0\\
                    0\quad \quad \quad \quad \quad \quad  \quad ,if\,{\bf{\bar Q}}_{ij}^{yl} < 0
                    \end{array} \right.
                    \end{equation}
                    \begin{equation}\label{eq12}
                    \frac{{\partial {\bf{Q}}_{ij}^{ll}}}{{\partial {{\bf{H}}_{*j}}}} = \left\{ \begin{array}{c}
                    \frac{{{{\bf{H}}_i}}}{{{{\left\| {{{\bf{H}}_i}} \right\|}_2} \cdot {{\left\| {{{\bf{H}}_j}} \right\|}_2}}}(\frac{{\partial {\bf{H}}_j^{\rm{T}}}}{{\partial {{\bf{H}}_{*j}}}}),if\,{\bf{\bar Q}}_{ij}^{ll} \ge 0\\
                    0\quad \quad \quad \quad \quad \quad \quad ,if\,{\bf{\bar Q}}_{ij}^{ll} < 0
                    \end{array} \right.
                    \end{equation}

                Then, we can calculate $\frac{{\partial J}}{{\partial {\theta _l}}}$ with $\frac{{\partial J}}{{\partial {{\bf{H}}_{*j}}}}$ by chain rule and update the parameter ${\theta _l}$ with BP algorithm.

              \textbf{Step 2:} Learn ${\theta _x}$, with ${{\bf{B}}^X}$, ${{\bf{B}}^Y}$, ${\theta _y}$ and ${\theta _l}$ fixed

                  When training ImgNet, we rewrite Eq.(7) as:
                      \begin{equation}\label{eq13}
                        \mathop {min}\limits_{{\theta _x}} J = \left\| {{{\bf{S}}^{xl}} - {{\bf{Q}}^{xl}}} \right\|_F^2 + \beta \left\| {{{\bf{S}}^{xx}} - {{\bf{Q}}^{xx}}} \right\|_F^2 + \gamma \left\| {{\bf{F}} - {{\bf{B}}^X}} \right\|_F^2
                      \end{equation}

                  Analogously, with ${{\bf{B}}^X}$, ${{\bf{B}}^Y}$, ${\theta _y}$ and ${\theta _l}$ fixed, we also use SGD and BP algorithm to learn the parameters ${\theta _x}$ of image hash function $f( \cdot )$. Firstly, the gradient is calculated by:
                    \begin{equation}\label{eq14}
                    \begin{array}{l}
                        \frac{{\partial J}}{{\partial {{\bf{F}}_{*i}}}} = 2({{\bf{Q}}^{xl}} - {{\bf{S}}^{xl}})\frac{{\partial {\bf{Q}}_{ij}^{xl}}}{{\partial {{\bf{F}}_{*i}}}}\\
                        \;\;\;\;\;\;\;\; + 2\beta ({{\bf{Q}}^{xx}} - {{\bf{S}}^{xx}})\frac{{\partial {\bf{Q}}_{ij}^{xx}}}{{\partial {{\bf{F}}_{*i}}}} + 2\gamma ({{\bf{F}}_{*i}} - {\bf{B}}_{*i}^X)
                    \end{array}
                    \end{equation}
                  where
                    \begin{equation}\label{eq15}
                    \frac{{\partial {\bf{Q}}_{ij}^{xl}}}{{\partial {{\bf{F}}_{*i}}}} = \left\{
                    \begin{array}{c}
                        \frac{1}{{{{\left\| {{{\bf{F}}_i}} \right\|}_2}}}(\frac{{\partial {{\bf{F}}_i}}}{{\partial {{\bf{F}}_{*i}}}}){(\frac{{{{\bf{H}}_j}}}{{{{\left\| {{{\bf{H}}_j}} \right\|}_2}}})^{\rm{T}}},if\,{\bf{\bar Q}}_{ij}^{xl} \ge 0\\
                        0\quad \quad \quad \quad \quad \quad \quad ,if\,{\bf{\bar Q}}_{ij}^{xl} < 0
                        \end{array} \right.
                    \end{equation}
                    \begin{equation}\label{eq16}
                    \frac{{\partial {\bf{Q}}_{ij}^{xx}}}{{\partial {{\bf{F}}_{*i}}}} = \left\{ \begin{array}{c}
                    \frac{1}{{{{\left\| {{{\bf{F}}_i}} \right\|}_2}}}(\frac{{\partial {{\bf{F}}_i}}}{{\partial {{\bf{F}}_{*i}}}}){(\frac{{{{\bf{F}}_j}}}{{{{\left\| {{{\bf{F}}_j}} \right\|}_2}}})^{\rm{T}}},if\,{\bf{\bar Q}}_{ij}^{xx} \ge 0\\
                    0\quad \quad \quad \quad \quad \quad \quad ,if\,{\bf{\bar Q}}_{ij}^{xx} < 0
                    \end{array} \right.
                    \end{equation}

                Then, we can calculate $\frac{{\partial J}}{{\partial {\theta _x}}}$ with $\frac{{\partial J}}{{\partial {{\bf{F}}_{*i}}}}$ by chain rule and update the parameter ${\theta _x}$ with BP algorithm.

              \textbf{Step 3}: Learn ${\theta _y}$, with ${{\bf{B}}^X}$, ${{\bf{B}}^Y}$, ${\theta _x}$ and ${\theta _l}$ fixed

                  Similarly, when training TxtNet, we rewrite Eq.(7) as follows:
                  \begin{equation}\label{17}
                  \mathop {min}\limits_{{\theta _y}} J = \alpha \left\| {{{\bf{S}}^{yl}} - {{\bf{Q}}^{yl}}} \right\|_F^2 + \beta \left\| {{{\bf{S}}^{yy}} - {{\bf{Q}}^{yy}}} \right\|_F^2 + \gamma \left\| {{\bf{G}} - {{\bf{B}}^Y}} \right\|_F^2
                  \end{equation}

                  Then with ${{\bf{B}}^X}$, ${{\bf{B}}^Y}$, ${\theta _x}$ and ${\theta _l}$ fixed, we compute the gradient:
                  \begin{equation}\label{18}
                  \begin{array}{l}
                    \frac{{\partial J}}{{\partial {{\bf{G}}_{*i}}}} = 2\alpha ({{\bf{Q}}^{yl}} - {{\bf{S}}^{yl}})\frac{{\partial {\bf{Q}}_{ij}^{yl}}}{{\partial {{\bf{G}}_{*i}}}}\\
                     + 2\beta ({{\bf{Q}}^{yy}} - {{\bf{S}}^{yy}}) \cdot \frac{{\partial {\bf{Q}}_{ij}^{yy}}}{{\partial {{\bf{G}}_{*i}}}} + 2\gamma ({{\bf{G}}_{*i}} - {\bf{B}}_{*i}^Y)
                  \end{array}
                  \end{equation}
                  where
                  \begin{equation}\label{eq19}
                  \frac{{\partial {\bf{Q}}_{ij}^{yl}}}{{\partial {{\bf{G}}_{*i}}}} = \left\{ \begin{array}{c}
                  \frac{1}{{{{\left\| {{{\bf{G}}_i}} \right\|}_2}}}(\frac{{\partial {{\bf{G}}_i}}}{{\partial {{\bf{G}}_{*i}}}}){(\frac{{{{\bf{H}}_j}}}{{{{\left\| {{{\bf{H}}_j}} \right\|}_2}}})^{\rm{T}}}\quad ,if\,{\bf{\bar Q}}_{ij}^{yl} \ge 0\\
                        0\quad \quad \quad \quad \quad ,if\,{\bf{\bar Q}}_{ij}^{yl} < 0
                  \end{array} \right.
                  \end{equation}

                  \begin{equation}\label{eq20}
                  \frac{{\partial {\bf{Q}}_{ij}^{yy}}}{{\partial {{\bf{G}}_{*i}}}} = \left\{ \begin{array}{c}
                  \frac{1}{{{{\left\| {{{\bf{G}}_i}} \right\|}_2}}}(\frac{{\partial {{\bf{G}}_i}}}{{\partial {{\bf{G}}_{*i}}}}){(\frac{{{{\bf{G}}_j}}}{{{{\left\| {{{\bf{G}}_j}} \right\|}_2}}})^{\rm{T}}}\quad ,if\,{\bf{\bar Q}}_{ij}^{yy} \ge 0\\
                        0\quad \quad \quad \quad \quad ,if\,{\bf{\bar Q}}_{ij}^{yy} < 0
                  \end{array} \right.
                  \end{equation}

                  Subsequently, we can calculate $\frac{{\partial J}}{{\partial {\theta _y}}}$ with $\frac{{\partial J}}{{\partial {{\bf{G}}_{*i}}}}$ by chain rule and update the parameter ${\theta _y}$ with BP algorithm.

              \textbf{Step 4:} Learn ${{\bf{B}}^X}$, with ${{\bf{B}}^Y}$, ${\theta _x}$, ${\theta _y}$ and ${\theta _l}$ fixed

                  After updating original image hash function, we learn the original hash codes ${{\bf{B}}^X}$ and rewrite Eq.(7) as:
                        \begin{equation}\label{eq21}
                          \begin{array}{c}
                            \mathop {\min }\limits_{{{\bf{B}}^X}} J = \mathop {\min }\limits_{{{\bf{B}}^X}} (\gamma \left\| {{\bf{F}} - {{\bf{B}}^X}} \right\|_F^2 + \frac{1}{2}\gamma \left\| {{\bf{H}} - {{\bf{B}}^X}} \right\|_F^2)\\
                             = \mathop {\min }\limits_{{{\bf{B}}^X}} tr\left( {{{( - 2\gamma {\bf{F}} - \gamma {\bf{H}})}^{\rm{T}}}({{\bf{B}}^X})} \right)\\
                            s.t. {\rm{   }}{{\bf{B}}^X} \in {\{  - 1, + 1\} ^{k \times m}}
                          \end{array}
                        \end{equation}

                  Obviously, when the sign of ${{\bf{B}}^X}$ is different from ${-(2\gamma {\bf{F}} + \gamma {\bf{H}})}$, Eq.21 will reach the minimum value. Namely, ${{\bf{B}}^X}$ and ${(2\gamma {\bf{F}} + \gamma {\bf{H})}}$ keep the same sign and formulates as:
                    \begin{equation}\label{eq22}
                      {{\bf{B}}^X} =  - sign( - (2\gamma {\bf{F}} + \gamma {\bf{H}})) = sign(2\gamma {\bf{F}} + \gamma {\bf{H}})
                    \end{equation}

                  \textbf{Step 5:} Learn ${{\bf{B}}^Y}$, with ${{\bf{B}}^X}$, ${\theta _x}$, ${\theta _y}$ and ${\theta _l}$ fixed

                      Similarly, after updating original text hash function, we have:
                        \begin{equation}\label{eq23}
                          {{\bf{B}}^Y} =  - sign( - (2\gamma {\bf{G}} + \gamma {\bf{H}})) = sign(2\gamma {\bf{G}} + \gamma {\bf{H}})
                        \end{equation}

                  Combing Eq.(8)-(23) and the Step 1-5 in original learning phase, the original hashing learning process can be summarized in Algorithm 1.

                    \begin{algorithm}[!t]
                        \scriptsize
                            \caption{The learning algorithm for original phase.}
                            \begin{algorithmic}[1]

                            \REQUIRE ~~\\ 
                            ${m}$ original data ${\bf{O}} = \{ {\bf{X,Y,L}}\} $; Code length ${k}$; Mini-batch size ${N_l} = {N_x} = {N_y} = 128$;
                            \ENSURE ~~\\ 
                            Original hash codes ${{\bf{B}}^X}$ and ${{\bf{B}}^Y}$; Network $h( \cdot )$, $f( \cdot )$ and $g( \cdot )$ with the parameters ${\theta _l}$, ${\theta _x}$ and ${\theta _y}$, respectively;

                            \STATE Initialize the parameters ${\theta _l}$, ${\theta _x}$ and ${\theta _y}$, hash code matrix ${{\bf{B}}^X}$ and ${{\bf{B}}^Y}$ randomly; iteration number ${T_l} = \lceil m/{N_l} \rceil$, ${T_x} = \lceil m/{N_x} \rceil$, ${T_y} = \lceil m/{N_y} \rceil$.
                            \STATE Calculate multi-label semantic similarity ${\bf{S}} = \{ {\bf{S}}^{xl},{\bf{S}}^{yl},{\bf{S}}^{ll},{\bf{S}}^{xx},{\bf{S}}^{yy}\} $ according to Eq.41, respectively.
                            \REPEAT
                            \FOR{$iter = 1$ to ${T_l}$}
                            \STATE Randomly sample ${N_l}$ instances from $\bf{L}$ to construct a mini-batch.
                            \STATE For each sampled instance ${l_i}$ in the mini-batch, calculate ${\bf{H}} = h({l_i};{\theta _l})$ by forward propagation.
                            \STATE Calculate the derivative according to Eq.(9).
                            \STATE Update the parameter ${\theta _l}$ by using back propagation.
                            \ENDFOR
                            \FOR{$iter = 1$ to ${T_x}$}
                            \STATE Randomly sample ${N_x}$ instances from $\bf{X}$ to construct a mini-batch.
                            \STATE For each sampled instance ${x_i}$ in the mini-batch, calculate ${\bf{F}} = f({x_i};{\theta _x})$ by forward propagation.
                            \STATE Calculate the derivative according to Eq.(14).
                            \STATE Update the parameter ${\theta _x}$ by using back propagation.
                            \ENDFOR
                            \FOR{$iter = 1$ to ${T_y}$}
                            \STATE Randomly sample ${N_y}$ instances from $\bf{Y}$ to construct a mini-batch.
                            \STATE For each sampled instance ${y_i}$ in the mini-batch, calculate ${\bf{G}} = g({y_i};{\theta _y})$ by forward propagation.
                            \STATE Calculate the derivative according to Eq.(18).
                            \STATE Update the parameter ${\theta _y}$ by using back propagation.
                            \ENDFOR
                            \STATE Update ${{\bf{B}}^X}$ according to Eq.(22).
                            \STATE Update ${{\bf{B}}^Y}$ according to Eq.(23).
                            \UNTIL a fixed number of iterations
                            \end{algorithmic}
                    \end{algorithm}

    \subsection{Lifelong Hash Codes Learning}
            \subsubsection{Lifelong Hashing Loss}
            To avoid performance degradation when training on incremental data and keep the learned hash function continuously usable, we design lifelong hashing loss which contains original similarity preserving loss and incremental similarity preserving loss. To this end, we firstly define original similarity preserving loss as Eq.(24) to keep the similarity between training and original data.
                \begin{equation}\label{eq24}
                  {J_{old}} = \left\| {{\rm{(}}{{\bf{B}}^X}} \right.{)^{\rm{T}}}{{\bf{A}}^X} - \left. {k{\bf{S}}^{to}} \right\|_F^{\rm{2}} + \left\| {{\rm{(}}{{\bf{B}}^Y}} \right.{)^{\rm{T}}}{{\bf{A}}^Y} - \left. {k{\bf{S}}^{to}} \right\|_F^{\rm{2}}
                \end{equation}
            where ${\bf{S}}^{to}$ with the size of ${m \times a}$ is the multi-label semantic similarity between training data and original data. Similarly, to keep the similarity between training and incremental data, we define incremental similarity preserving loss as:
                \begin{equation}\label{eq25}
                  {J_{new}} = \left\| {{\rm{(}}{{{\bf{B}}}^{X'}}} \right.{)^{\rm{T}}}{{\bf{A}}^X} - \left. {k{\bf{S}}^{ti}} \right\|_F^{\rm{2}} + \left\| {{\rm{(}}{{\bf{B}}^{Y'}}} \right.{)^{\rm{T}}}{{\bf{A}}^Y} - \left. {k{\bf{S}}^{ti}} \right\|_F^{\rm{2}},
                \end{equation}
            where ${\bf{S}}^{ti}$ with the size of ${n \times a}$ is multi-label semantic similarity between training and incremental data. To approximate discrete hash codes, we formulate quantization loss as:
                \begin{equation}\label{eq26}
                  {J_{quan}} = \left\| {{{\bf{B}}^{X'}}} \right. - \left. {\bf{F}'} \right\|_F^{\rm{2}} + \left\| {{{\bf{B}}^{Y'}}} \right. - \left. {\bf{G}'} \right\|_F^{\rm{2}}
                \end{equation}
            where ${{\bf{B}}^{X'}} \in {\{  - 1, + 1\} ^{k \times {a_2}}}$ and ${{\bf{B}}^{Y'}} \in {\{  - 1, + 1\} ^{k \times {a_2}}}$ are the learned binary hash codes of ${a_2}$ incremental data sampled to training set. ${\bf{F}}' = f{\rm{(}}{x_a}{\rm{;}}{\theta _x}{\rm{)}} \in {R^{k \times {a_2}}}$ and ${\bf{G}}' = g{\rm{(}}{y_a}{\rm{;}}{\theta _y}{\rm{)}} \in {R^{k \times {a_2}}}$ are outputs of training data sample from incremental set through the updated hash functions.

            Considering that the larger is the information entropy, the larger is information amount obtained after eliminating uncertainty, we introduce balance loss which balances the number of -1 and +1 for each bit to maximize the entropy of hash bits as:
                \begin{equation}\label{eq27}
                  {J_{balance}} = \left\| {\bf{F}'} \right. \cdot \left. 1 \right\|_F^{\rm{2}} + \left\| \bf{G'} \right. \cdot \left. 1 \right\|_F^{\rm{2}}
                \end{equation}

            By incorporating Eq.(27), the accumulation of each row of $\bf{F}'$ and $\bf{G}'$ is as close to 0 as possible, which achieves balanced number of -1 and +1 on each bit. Combining original similarity preserving loss, incremental similarity preserving loss, quantization loss and balance loss, the final objective function of lifelong hash learning phase can be defined as:
                \begin{equation}\label{eq28}
                  \begin{array}{c}
                    \mathop {\min }\limits_{{{\bf{B}}^{X'}},{{\bf{B}}^{Y'}},{\theta _x},{\theta _y}} J = {J_{old}} + {J_{new}} + \lambda {J_{quan}} + \mu {J_{balance}}\\
                    s.t.\quad {{\bf{B}}^{X'}} \in {\{  - 1, + 1\} ^{k \times n}}\\
                    {{\bf{B}}^{Y'}} \in {\{  - 1, + 1\} ^{k \times n}}
                    \end{array}
                \end{equation}
            where $\lambda$ and $\mu$ are hyper-parameters that control the weight of each part.

            \subsubsection{Optimization}
                We also utilize the alternating learning strategy to update the hash function $f( \cdot )$ and $g( \cdot )$, and their respective parameters ${\theta _x}$ and ${\theta _y}$, so as to find ${{\bf{B}}^{X'}}$ and ${{\bf{B}}^{Y'}}$.

            \textbf{Step 1:} Learn ${\theta _x}$, with ${{\bf{B}}^{X'}}$, ${{\bf{B}}^{Y'}}$, ${\theta _y}$ fixed

                Similar to the optimization of the original learning phase, with ${{\bf{B}}^{X'}}$, ${{\bf{B}}^{Y'}}$ and ${\theta _y}$ fixed, SGD and BP algorithm are utilized to update the parameter ${\theta _x}$ of image hash function $f( \cdot )$. Note that continuous function $tanh( \cdot )$ is used to fit ${{\bf{A}}^X}$, and Eq.(28) can be rewritten as:
                    \begin{equation}\label{eq29}
                      \begin{array}{l}
                          \mathop {{\rm{min}}}\limits_{{\theta _x}} J = \left\| {{\rm{(}}{{\bf{B}}^X}} \right.{)^{\rm{T}}}{{\bf{A}}^X} - \left. {k{\bf{S}}^{to}} \right\|_F^{\rm{2}} + \left\| {{\rm{(}}{{\bf{B}}^{X'}}} \right.{)^{\rm{T}}}{{\bf{A}}^X} - \left. {k{\bf{S}}^{ti}} \right\|_F^{\rm{2}} \\
                          \;\;\;\;\;\;\;\;\;\; + \lambda \left\| {{{\bf{B}}^{X'}} - {\bf{F}}'} \right\|_F^{\rm{2}} + \mu \left\| {{\bf{F}}' \cdot {\bf{1}}} \right\|_F^{\rm{2}}
                      \end{array}
                    \end{equation}

                For each image feature ${x_i}$ in training set, we compute the gradient as:
                    \begin{equation}\label{eq30}
                      \begin{array}{l}
                      \frac{{\partial J}}{{\partial {\bf{F}}'_{*i}}} = 2({({\bf{B}}_{*i}^X)^{\rm{T}}}{\bf{A}}_{*i}^X{\bf{B}}_{*i}^X) + 2({({\bf{B}}_{*i}^{X'})^{\rm{T}}}{\bf{A}}_{*i}^X{\bf{B}}_{*i}^{X'}) \\
                      \;\;\;\;\;\;\;\; + 2\; \lambda ({\bf{F}}'_{*i} - {\bf{B}}_{*i}^{X'}) + 2\mu {\bf{F}}' \cdot {\bf{1}}
                      \end{array}
                    \end{equation}

                Then, we can compute $\frac{{\partial J}}{{\partial {\theta _x}}}$ with $\frac{{\partial J}}{{\partial {{\bf{F}}'_{*i}}}}$ by chain rule and update the parameter ${\theta _x}$ with BP algorithm.

            \textbf{Step 2:} Learn ${\theta _y}$, with ${{\bf{B}}^{X'}}$, ${{\bf{B}}^{Y'}}$, ${\theta _x}$ fixed

                 We also use $tanh( \cdot )$ to fit ${{\bf{A}}^Y}$ to update parameter ${\theta _y}$ of text hash function $g( \cdot )$ with ${{\bf{B}}^{X'}}$, ${{\bf{B}}^{Y'}}$, ${\theta _x}$ fixed. Then, we rewrite Eq.(28) as:
                    \begin{equation}\label{eq31}
                      \begin{array}{c}
                      \mathop {{\rm{min}}}\limits_{{\theta _y}} J = \left\| {{\rm{(}}{{\bf{B}}^Y}} \right.{)^{\rm{T}}}{{\bf{A}}^Y} - \left. {k{\bf{S}}^{to}} \right\|_F^{\rm{2}} + \left\| {{\rm{(}}{{\bf{B}}^{Y'}}} \right.{)^{\rm{T}}}{{\bf{A}}^Y} - \left. {k{\bf{S}}^{ti}} \right\|_F^{\rm{2}} \\
                      + \lambda \left\| {{{\bf{B}}^{Y'}} - {\bf{G}}'} \right\|_F^{\rm{2}} + \mu \left\| {{\bf{G}}' \cdot {\bf{1}}} \right\|_F^{\rm{2}}
                      \end{array}
                    \end{equation}

                For each text feature ${y_j}$ in training set, we compute the gradient:
                    \begin{equation}\label{eq32}
                      \begin{array}{c}
                          \frac{{\partial J}}{{\partial {\bf{G}}'_{*j}}} = 2({({\bf{B}}_{*j}^Y)^{\rm{T}}}{\bf{A}}_{*j}^Y{\bf{B}}_{*j}^Y) + 2({({\bf{B}}_{*j}^{Y'})^{\rm{T}}}{\bf{A}}_{*j}^Y{\bf{B}}_{*j}^{Y'})\\
                          + 2\lambda ({\bf{G}}'_{*j} - {\bf{B}}_{*j}^{Y'}) + 2\mu {\bf{G}}' \cdot {\bf{1}}
                      \end{array}
                    \end{equation}

                Then, we can compute $\frac{{\partial J}}{{\partial {\theta _y}}}$ with $\frac{{\partial J}}{{\partial {{\bf{G}}'_{*i}}}}$ by chain rule and update the parameter ${\theta _y}$ with BP algorithm.

            \textbf{Step 3:} Learn ${{\bf{B}}^{X'}}$, with ${{\bf{B}}^{Y'}}$, ${\theta _y}$, ${\theta _y}$ fixed

                When ${{\bf{B}}^{Y'}}$, ${\theta _y}$, ${\theta _y}$ are fixed, we denote ${{\bf{B}}^{X'}} \in {\{  - 1, + 1\} ^{k \times {a_2}}}$ in Eq.(26) as ${\bf{B}}_{{a_2}}^{X'}$. So we have:
                    \begin{equation}\label{eq33}
                      \begin{array}{c}
                        \mathop {\min }\limits_{{{\bf{B}}^{X'}}} J = \left\| {{{{\rm{(}}{{\bf{B}}^{X'}})}^{\rm{T}}}{{\bf{A}}^X} - k{\bf{S}}^{ti}} \right\|_F^2 + \lambda \left\| {{\bf{B}}_{{a_2}}^{X'} - {\bf{F}}'} \right\|_F^2\\
                         = \left\| {{{{\rm{(}}{{\bf{B}}^{X'}})}^{\rm{T}}}{{\bf{A}}^X}} \right\|_F^2 - 2ktr({{\bf{B}}^{X'}}{\bf{S}}^{ti}{({{\bf{A}}^X})^{\rm{T}}}) + {k^2}\left\| {\bf{S}}^{ti} \right\|_F^2\\
                         + \lambda {\rm{(}}\left\| {{\bf{B}}_{{a_2}}^{X'}} \right\|_F^2 - 2tr({\bf{B}}_{{a_2}}^{X'}{\bf{F}}{'^{\rm{T}}}) + \left\| {{\bf{F}}'} \right\|_F^2{\rm{)}}
                    \end{array}
                    \end{equation}

                The goal is to update ${{\bf{B}}^{X'}}$, and Eq.(33) can be rewritten as:
                    \begin{equation}\label{eq34}
                      \begin{array}{c}
                      \mathop {\min }\limits_{{{\bf{B}}^{X'}}} J = \left\| {{{{\rm{(}}{{\bf{B}}^{X'}})}^{\rm{T}}}{{\bf{A}}^X}} \right\|_F^2 - 2ktr({{\bf{B}}^{X'}}{\bf{S}}^{ti}{({{\bf{A}}^X})^{\rm{T}}}) \\
                      - 2\lambda tr({\bf{B}}_{{a_2}}^{X'}{\bf{F}}{'^{\rm{T}}})
                      \end{array}
                    \end{equation}

                To further simplify Eq.(34), ${\bf{B}}_{{a_2}}^{X'} \in {\{  - 1, + 1\} ^{k \times {a_2}}}$ in the third term can be transformed to ${{\bf{B}}^{X'}} \in {\{  - 1, + 1\} ^{k \times n}}$. Thus, we define a $k \times n$ matrix ${\bf{\tilde F}}'$ as an extension of ${\bf{F}}'$, i.e., the corresponding bits in ${\bf{\tilde F}}'$ retain hash codes in ${\bf{F}}'$ and set the other bits to 0. In this case, Eq.(34) can be rewritten as:
                    \begin{equation}\label{eq35}
                      \begin{array}{c}
                        \mathop {\min }\limits_{{{\bf{B}}^{X'}}} J = \left\| {{{{\rm{(}}{{\bf{B}}^{X'}})}^{\rm{T}}}{{\bf{A}}^X}} \right\|_F^2 - 2ktr({{\bf{B}}^{X'}}{\bf{S}}^{ti}{({{\bf{A}}^X})^{\rm{T}}}) \\- 2\lambda tr({{\bf{B}}^{X'}}{\bf{\tilde F}}{'^{\rm{T}}})\\
                         = \left\| {{{{\rm{(}}{{\bf{B}}^{X'}})}^{\rm{T}}}{{\bf{A}}^X}} \right\|_F^2 - 2tr({({{\bf{B}}^{X'}})^{\rm{T}}}(k{{\bf{A}}^X}{({\bf{S}}^{ti})^{\rm{T}}} + \lambda {\bf{\tilde F}}'))\\
                         = \left\| {{{{\rm{(}}{{\bf{B}}^{X'}})}^{\rm{T}}}{{\bf{A}}^X}} \right\|_F^2 + tr({({{\bf{B}}^{X'}})^{\rm{T}}}{{\bf{P}}^X})
                      \end{array}
                    \end{equation}
                where ${{\bf{P}}^X} =  - 2k{{\bf{A}}^X}{({\bf{S}}^{ti})^{\rm{T}}} - {\rm{2}}\lambda {\bf{\tilde F}}'$.

                To ensure retrieval accuracy, we do not adopt the relaxation strategy, but use discrete cyclic coordinate descent (DCC) algorithm to optimize Eq.(35) and update ${{\bf{B}}^{X'}}$ bit by bit. For this, we denote ${\bf{B}}_{r*}^{X'}$ as the $\emph{r}$-th row of ${{\bf{B}}^{X'}}$ and ${\bf{B}}_{ - r}^{X'}$ as the matrix of ${{\bf{B}}^{X'}}$ excluding ${\bf{B}}_{r*}^{X'}$. For ${{\bf{A}}^X}$ and ${{\bf{P}}^X}$, ${\bf{A}}_{r*}^{X}$ is the ${r}$-th row of ${{\bf{A}}^X}$, ${\bf{A}}_{ - r}^{X}$ is the matrix of ${\bf{A}}_{r*}^{X}$ excluding ${\bf{A}}_{r*}^{X}$, and ${\bf{P}}_{r*}^{X}$ is the ${r}$-th row of ${{\bf{P}}^X}$. Hence, Eq.(35) is transformed as follows:
                    \begin{equation}\label{eq36}
                      \begin{array}{c}
                        \mathop {\min }\limits_{{\bf{B}}_{r*}^{X'}} J = \left\| {{{{\rm{(}}{{\bf{B}}^{X'}})}^{\rm{T}}}{{\bf{A}}^X}} \right\|_F^2 + tr({({{\bf{B}}^{X'}})^{\rm{T}}}{{\bf{P}}^X})\\
                         = tr({\bf{B}}_{r*}^{X'}(2{({\bf{B}}_{ - r}^{X'})^{\rm{T}}}{\bf{A}}_{ - r}^X{({\bf{A}}_{r*}^X)^{\rm{T}}} + {({\bf{P}}_{r*}^X)^{\rm{T}}}))
                        \end{array}
                    \end{equation}

                It is easy to find that Eq.(36) achieves the minimum when each bit in ${\bf{B}}_{r*}^{X'}$ has opposite sign to the corresponding bit in $\left( {2{{({\bf{B}}_{ - r}^{X'})}^{\rm{T}}}{\bf{A}}_{ - r}^X{{({\bf{A}}_{r*}^X)}^{\rm{T}}} + {{({\bf{P}}_{r*}^X)}^{\rm{T}}}} \right)$. Therefore, the optimal solution of Eq.(36) is
                    \begin{equation}\label{eq37}
                      {\bf{B}}_{r*}^{X'} =  - sign(2{({\bf{B}}_{ - r}^{X'})^{\rm{T}}}{\bf{A}}_{ - r}^X{({\bf{A}}_{r*}^X)^{\rm{T}}} + {({\bf{P}}_{r*}^X)^{\rm{T}}})
                    \end{equation}

                We can compute the ${\emph{r}}$-th row of ${{\bf{B}}^{X'}}$ with the above formula, and then update all row of ${{\bf{B}}^{X'}}$ by replacing ${\bf{B}}_{r*}^{X'}$.

            \textbf{Step 4:} Learn ${{\bf{B}}^{Y'}}$, with ${{\bf{B}}^{X'}}$, ${\theta _y}$, ${\theta _y}$ fixed

                Similarly, when ${{\bf{B}}^{X'}}$, ${\theta _y}$ and ${\theta _y}$ are fixed, we simplify Eq.(28) as follows:
                    \begin{equation}\label{eq38}
                      \mathop {\min }\limits_{{{\bf{B}}^{Y'}}} J = \left\| {{{{\rm{(}}{{\bf{B}}^{Y'}})}^{\rm{T}}}{{\bf{A}}^Y}} \right\|_F^2 + tr({{\rm{(}}{{\bf{B}}^{Y'}})^{\rm{T}}}{{\bf{P}}^Y}),
                    \end{equation}
                where ${{\bf{P}}^Y} =  - 2k{{\bf{A}}^Y}({\bf{S}}^{ti}){^{\rm{T}}} - {\rm{2}}\lambda {\bf{\tilde G}}'$. ${\bf{\tilde G}}'$ is a $k \times n$ extension matrix of ${\bf{G}}'$, i.e., the corresponding bits in ${\bf{\tilde G}}'$ retain hash codes in ${\bf{G}}'$ and the other bits are 0. Let ${\bf{B}}_{r*}^{Y'}$ be the $\emph{r}$-th row of ${{\bf{B}}^{Y'}}$, ${\bf{B}}_{ - r}^{Y'}$ be the matrix of ${{\bf{B}}^{Y'}}$ excluding ${\bf{B}}_{r*}^{Y'}$, ${\bf{A}}_{r*}^{Y}$ be the $\emph{r}$-th row of ${{\bf{A}}^{Y}}$, ${\bf{A}}_{ - r}^{Y}$ be the matrix of ${{\bf{A}}^{Y}}$ excluding ${\bf{A}}_{r*}^{Y}$, ${\bf{P}}_{r*}^{Y}$ be the $\emph{r}$-th row of ${{\bf{P}}^{Y}}$. Thus, Eq.(38) is converted to the following formula:
                    \begin{equation}\label{eq39}
                      \begin{array}{c}
                        \mathop {\min }\limits_{{\bf{B}}_{r*}^{Y'}} J = \left\| {{{{\rm{(}}{{\bf{B}}^{Y'}})}^{\rm{T}}}{\bf{A}}^Y} \right\|_F^2 + tr({({{\bf{B}}^{Y'}})^{\rm{T}}}{{\bf{P}}^Y})\\
                         = tr({\bf{B}}_{r*}^{Y'}(2{({\bf{B}}_{ - r}^{Y'})^{\rm{T}}}{\bf{A}}_{ - r}^Y{({\bf{A}}_{r*}^Y)^{\rm{T}}} + {({\bf{P}}_{r*}^Y)^{\rm{T}}}))
                        \end{array}
                    \end{equation}

                It is easy to find that ${\bf{B}}_{r*}^{Y'}$ has opposite sign to the corresponding bit in $\left( {2{{({\bf{B}}_{ - r}^{Y'})}^{\rm{T}}}{\bf{A}}_{ - r}^Y{{({\bf{A}}_{r*}^Y)}^{\rm{T}}} + {{({\bf{P}}_{r*}^Y)}^{\rm{T}}}} \right)$. Hence, we can obtain the optimal solution of Eq.(39), that is
                    \begin{equation}\label{eq40}
                      {\bf{B}}_{r*}^{Y'} =  - sign(2{({\bf{B}}_{ - r}^{Y'})^{\rm{T}}}{\bf{A}}_{ - r}^Y{({\bf{A}}_{r*}^Y)^{\rm{T}}} + {({\bf{P}}_{r*}^Y)^{\rm{T}}})
                    \end{equation}

                Subsequently, the $\emph{r}$-th row of ${{\bf{B}}^{X'}}$ can be calculated with the above formula, and all row of ${{\bf{B}}^{X'}}$ will be updated by replacing ${\bf{B}}_{r*}^{X'}$.

                Combing Eq.(24)-(40) and the Step 1-4 in lifelong learning phase, the lifelong hashing learning process can be summarized in Algorithm 2.
                    \begin{algorithm}[!t]
                    \scriptsize
                        \caption{The learning algorithm for lifelong phase.}
                        \begin{algorithmic}[1]

                        \REQUIRE ~~\\ 
                        ${n}$ incremental data ${\bf{O}}' = \{ {\bf{X}}',{\bf{Y}}',{\bf{L}}'\} $; Original hash codes ${{\bf{B}}^X}$ and ${{\bf{B}}^Y}$; Number of training data ${a}$; Code length ${k}$; Mini-batch size ${N_x} = {N_y} = 128$.
                        \ENSURE ~~\\ 
                        Incremental hash codes ${{\bf{B}}^{X'}}$ and ${{\bf{B}}^{Y'}}$; Updated parameters ${\theta _x}$ and ${\theta _y}$.

                        \STATE Sample ${a}$ data from original set and incremental set randomly as training set ${{\bf{O}}_a} = \{ {{\bf{X}}_a}{\bf{,}}{{\bf{Y}}_a}{\bf{,}}{{\bf{L}}_a}\} $.
                        \STATE Initialize parameters ${\theta _x}$ and ${\theta _y}$, hash code matrix ${{\bf{B}}^{X'}}$ and ${{\bf{B}}^{Y'}}$ randomly; iteration number ${T_x} = \lceil a/{N_x} \rceil$, ${T_y} = \lceil a/{N_y} \rceil$.
                        \STATE Calculate multi-label semantic similarity ${\bf{S}} = \{{{\bf{S}}^{to}}, {{\bf{S}}^{ti}}\}$ according to Eq.(41).
                        \REPEAT
                        \FOR{$iter = 1$ to ${T_x}$}
                        \STATE Randomly sample ${N_x}$ instances from ${{\bf{X}}_a}$ to construct a mini-batch.
                        \STATE For each sampled instance ${x_i}$ in the mini-batch, calculate ${\bf{F}}' = f({x_i};{\theta _x})$ by forward propagation.
                        \STATE Calculate the derivative according to Eq.(30).
                        \STATE Update the parameter ${\theta _x}$ by using back propagation.
                        \ENDFOR
                        \FOR{$iter = 1$ to ${T_y}$}
                        \STATE Randomly sample ${N_y}$ instances from ${{\bf{Y}}_a}$ to construct a mini-batch.
                        \STATE For each sampled instance ${y_i}$ in the mini-batch, calculate ${\bf{G}}' = g({y_i};{\theta _y})$ by forward propagation.
                        \STATE Calculate the derivative according to Eq.(32).
                        \STATE Update the parameter ${\theta _y}$ by using back propagation.
                        \ENDFOR
                        \STATE Update ${{\bf{B}}^{X'}}$ according to Eq.(37).
                        \STATE Update ${{\bf{B}}^{Y'}}$ according to Eq.(40).
                        \UNTIL a fixed number of iterations
                        \end{algorithmic}
                \end{algorithm}

    \subsection{Multi-label Semantic Similarity}
        As mentioned before, the existing online cross-modal hashing methods utilize single-label similarity to evaluate correlation of two instances, i.e., they are considered as similar as long as two instances have one common label. For example, when instance $x_1$ shares one common label with instance $x_2$, and shares three same labels with instance $x_3$, the single-label concept holds that the semantic similarity of $x_1$ and $x_2$, $x_1$ and $x_3$ are both 1. It is obvious that instance $x_1$ and $x_3$ are more similar as they have more common labels, which illustrates the limitation of single-label similarity in describing similarity relationships. Thus, to make full use of label information, we adopt multi-label semantic similarity instead of coarse-grained single-label similarity, which is defined as follows:
            \begin{equation}\label{eq41}
              {\bf{S}}_{ij}^{{o_1}{o_2}} = \frac{{{\bf{L}}_i^{{o_1}}}}{{{{\left\| {{\bf{L}}_i^{{o_1}}} \right\|}_2}}} \cdot {(\frac{{{\bf{L}}_j^{{o_2}}}}{{{{\left\| {{\bf{L}}_j^{{o_2}}} \right\|}_2}}})^{\rm{T}}}
            \end{equation}
        where ${\bf{L}}_i^{{o_1}}$ denotes the label of $i$-th instance in ${o_1}$ modality , ${\bf{L}}_j^{{o_2}}$ denotes the label of $j$-th instance in ${o_2}$ modality, and ${o_{1,2}} \in \{ x,y,l\} $. The range of ${\bf{S}}_{ij}^{{o_1}{o_2}}$ is $[0,1]$. If instances ${o_i}$ and ${o_j}$ have a larger ${\bf{S}}_{ij}^{{o_1}{o_2}}$, it means that they are more semantically similar, otherwise less similar.

        For instances ${o_i}$ and ${o_j}$ and their corresponding hash codes ${b_i}$ and ${b_j}$, we utilize Hamming distance $di{s_H}({b_i},{b_j}) = \frac{1}{2}(K -  < {b_i},{b_j} > )$ to measure the similarity between two hash codes, where $ < {b_i},{b_j} > $ denotes the inner product of ${b_i}$ and ${b_j}$. When using single-label semantic similarity ${\bf{S}}{'_{ij}} = {l_i} \cdot {({l_j})^{\rm{T}}}$, the likelihood function can be written as:
            \begin{equation}\label{eq42}
            \begin{array}{l}
                P({\bf{S}}{'_{ij}}|{b_i},{b_j}) = \left\{ \begin{array}{l}
                {\rm{\sigma }}( < {b_i},{b_j} > )\quad {\kern 1pt} \,{\kern 1pt} ,{\bf{S}}{'_{ij}} = 1\\
                1 - {\rm{\sigma }}( < {b_i},{b_j} > ),{\bf{S}}{'_{ij}} = 0
                \end{array} \right.\\
                 = {\bf{S}}{'_{ij}}{\rm{\sigma }}( < {b_i},{b_j} > ) + (1 - {\bf{S}}{'_{ij}})(1 - {\rm{\sigma }}( < {b_i},{b_j} > )),
                \end{array}
            \end{equation}
        where ${\rm{\sigma }}( < {b_i},{b_j} > ) = \frac{1}{{1 + {e^{ -  < {b_i},{b_j} > }}}}$. Given hash codes ${b_i}$ and ${b_j}$, the probability of ${\bf{S}}'$ is:
            \begin{equation}\label{eq43}
            {L_1} = P({\bf{S}}{'_{ij}}|{b_i},{b_j}) = \prod\limits_{i,j = 1}^n {\frac{{{e^{({\bf{S}}{'_{ij}} - 1) < {b_i},{b_j} > }}}}{{1 + {e^{ -  < {b_i},{b_j} > }}}}}
            \end{equation}

        To facilitate calculation, take the logarithm of Eq.(43):
            \begin{equation}\label{eq44}
            \begin{array}{l}
            \log {L_1} = \log (P({\bf{S}}{'_{ij}}|{b_i},{b_j}))\\
             = \sum\limits_{i,j = 1}^n {[({l_i} \cdot {{({l_j})}^{\rm{T}}}) < {b_i},{b_j} >  - \log (1 + {e^{ < {b_i},{b_j} > }})]}
            \end{array}
            \end{equation}

        Similarly, for multi-label similarity ${{\bf{S}}_{ij}} = \frac{{{l_i}}}{{{{\left\| {{l_i}} \right\|}_2}}} \cdot {(\frac{{{l_j}}}{{{{\left\| {{l_j}} \right\|}_2}}})^{\rm{T}}}$, the likelihood function is:
            \begin{equation}\label{eq45}
                P({{\bf{S}}_{ij}}|{b_i},{b_j}) = {{\bf{S}}_{ij}}{\rm{\sigma }}( < {b_i},{b_j} > )
            \end{equation}

        Given hash codes ${b_i}$ and ${b_j}$, the probability of ${\bf{S}}$ is:
            \begin{equation}\label{eq46}
                {L_2} = P({{\bf{S}}_{ij}}|{b_i},{b_j}) = \prod\limits_{i,j = 1}^n {\frac{{{{\bf{S}}_{ij}}}}{{1 + {e^{ -  < {b_i},{b_j} > }}}}}
            \end{equation}

        Take the logarithm of the above formula:
            \begin{equation}\label{eq47}
            \begin{array}{l}
                \log {L_2} = \log (P({{\bf{S}}_{ij}}|{b_i},{b_j}))\\
                 = \sum\limits_{i,j = 1}^n {[\log \frac{{{l_i}}}{{{{\left\| {{l_i}} \right\|}_2}}} \cdot {{(\frac{{{l_j}}}{{{{\left\| {{l_j}} \right\|}_2}}})}^{\rm{T}}} +  < {b_i},{b_j} >  - \log (1 + {e^{ < {b_i},{b_j} > }})]}
            \end{array}
            \end{equation}

        Combining Eq.(44) and Eq.(47), we can see both ${L_1}$ and ${L_2}$ generate the same results when query instance ${o_i}$ and retrieved instance ${o_j}$ have only one common label. When ${o_i}$ and ${o_j}$ have more than one common label which is more common for existing cross-modal datasets, optimizing ${L_2}$ yields larger inner product and smaller Hamming distance.

        Given fixed Hamming radius, query instance ${o_i}$, retrieved instance ${o_j}$ and their corresponding hash codes ${b_i}$ and ${b_j}$, the conditional probability of relevance $rel({o_i},{o_j})$ can be defined by the Bernoulli distribution as:
            \begin{equation}\label{eq48}
            P(rel({o_i},{o_j})|{b_i},{b_j}) = \left\{ \begin{array}{l}
            \delta (di{s_H}({b_i},{b_j}))\quad {\kern 1pt} \,{\kern 1pt} ,rel({o_{ij}}) = 1\\
            1 - \delta (di{s_H}({b_i},{b_j})),rel({o_{ij}}) = 0
            \end{array} \right.
            \end{equation}

        From Eq.(48) we can see that smaller Hamming distance $di{s_H}({b_i},{b_j})$ makes larger conditional probability $P(rel({o_i},{o_j}) = 1|{b_i},{b_j})$, which signifies that ${o_i}$ and ${o_j}$ should be judged as relevant. Otherwise, larger conditional probability $P(rel({o_i},{o_j}) = 0|{b_i},{b_j})$ indicates that it should be judged as irrelevant. Therefore, for instances ${o_i}$ and ${o_j}$, it is known that the Hamming distance when optimizing ${L_2}$ is smaller than it when optimizing ${L_1}$, namely, the probability that ${o_i}$ and ${o_j}$ are judged to be relevant when optimizing ${L_2}$ is larger than the probability when optimizing ${L_1}$.

        Hash lookup, a widely used retrieval protocol in hashing-based retrieval\cite{10}, considers the retrieved instance whose Hamming distance to the query is less than Hamming radius as a positive sample. When measuring the precision of hash lookup protocol, we hope that a good hashing method can retrieve as many positive samples as possible, i.e., when the query instance ${o_i}$ and the retrieved instance ${o_j}$ are true relevant, the probability of being judged as relevant should be as large as possible, denoted as:
            \begin{equation}\label{eq49}
                Precision({o_i},{o_j}) \propto P(rel({o_i},{o_j}) = 1|{b_i},{b_j})
            \end{equation}

        As shown in Eq.(49), multi-label semantic similarity enables smaller Hamming distance between relevant query instances ${o_i}$ and retrieved instances ${o_j}$, which results in higher precision. Thus, we can conclude that the proposed multi-label semantic similarity can effectively improve retrieved precision, which will be proved with the subsequent experimental results.

    \subsection{Complexity Analysis}
        The computational complexity of original learning phase is mainly generated by optimizing neural network parameters and learning original hash codes. Specifically, the complexity of optimizing parameters ${\theta _l}$, ${\theta _x}$ and ${\theta _y}$ of LabelNet, ImgNet and TxtNet can be calculated by Eq.(9), Eq.(14) and Eq.(18), respectively, as $O({m^2}k)$. The complexity of learning original hash codes is found by Eq.(22) and Eq.(23), as $O(mk \times mk) = O({m^2}{k^2})$. Due to $k \ll m$, the computational complexity of original learning is $O({m^2})$.

        The computational complexity of lifelong learning phase is mainly generated by updating parameters of hash function $f( \cdot )$, $g( \cdot )$ and learning incremental hash codes. Specifically, the complexity of updating parameter ${\theta _x}$ and ${\theta _y}$ can be calculated by Eq.(30) and Eq.(32), as $O((m + n)ak)$, while the complexity of updating the incremental hash codes is computed by Eq.(37) and Eq.(40), as $O(nk \times ak) = O(an{k^2})$. Since ${a}$ and ${k}$ are much smaller than ${m}$ and ${n}$, the computational complexity of updating the parameters is $O(m + n)$, and that of learning incremental hash codes is $O(n)$.

\section{Experiment}
    To verify the effectiveness of DLCH, extensive experiments are carried out on three widely-used benchmark datasets. We firstly compare our DLCH with several deep cross-modal hashing algorithms including non-continuous and online methods in retrieval performance, and analyze the results. Then, discuss DLCH and its variants to analyze function of composition. The implementation based on PyTorch can be available at https://github.com.

    \subsection{Datasets}
        \textbf{MIRFlickr25K} is a multi-label dataset which contains about 25000 image-text pairs associated with 24 class labels and are collected from Flickr. The text annotation for each point is represented as 1386-dimensional bag-of-words vector. In our experiment, we select the image-text pairs with at least 20 text annotations, which yields 20,015 image text pairs.

        \textbf{NUS-WIDE} is another common-used multi-label dataset which includes over 269,000 image-text pairs with 81 class labels. The text annotation for each data point is represented as 1000-dimensional bag-of-words vector. In our experiment, we choose the pairs which belong to the 21 most frequently used labels, where each label contains at least 5,000 data, resulting in over 195,000 image-text pairs.

        \textbf{Wiki} is a single-label dataset which contains 2,866 image-text pairs with 10 class labels. The text annotation for each data point is represented as a 128-dimensional SIFT vector. In our experiment, we choose all the image-text pairs.
            \begin{table}[!t]
                \caption{\small{Detailed settings of experimental datasets}}
        		\label{tab2}
                \centering
                \begin{tabular}{|c|c|c|c|}
                \hline
                    Set                                      & \multicolumn{1}{c|}{MIRFlickr} & \multicolumn{1}{c|}{NUS-WIDE}    & \multicolumn{1}{c|}{Wiki} \\
                \hline\hline
                	Total                                    & 20015                             & 195834                           & 2866                   \\ \hline
                    Query                                    & 2000                              & 2100                             & 693                    \\ \hline
                    Retrieval                                & 18015                             & 193734                           & 2173                   \\
                \hline\hline
                                                             & 23/1                              & 20/1                             & 9/1                    \\ \cline{2-4}
                    \multirow{1}*{Class of original set}     & 22/2                              & 19/2                             & 8/2                    \\ \cline{2-4}
                    \multirow{1}*{VS}                        & 21/3                              & 18/3                             & 7/3                    \\ \cline{2-4}
                    \multirow{1}*{Class of incremental set}  & 20/4                              & 17/4                             & 6/4                    \\ \cline{2-4}
                                                             & 12/12                             & 10/11                            & 4/6                    \\ \hline
                \end{tabular}
            \end{table}

        For all datasets, 10\% pairs from each class will be chose to form testing set and the rest are database set. Within database set, 90\% pairs from will be used as training set and the remaining pairs are used for validation set. We then resize all of the images to be the size of 256$\times$256. To simulate that incremental data with categories appears continuously, we divide the benchmark datasets into two parts, i.e., original set and incremental set, according to class labels. For WiKi, the split setting '9/1' denotes that nine classes are selected as original set, and the remain one is used for incremental set. For MIRFlickr, '23/1' means that all the data in original set contain at most 23 class labels, and the rest one class data is treated as incremental set. Similar setting for NUS-WIDE, and the same are the other settings. The statistics and split setting of three benchmark datasets are shown in TABLE II.
                   \begin{table*}[!t]
            \caption{\small{MAP Scores of Cross-modal Retrieval Task on Benchmark Datasets with Different Lengths of Hash Codes. The \textbf{bold} and \underline{underlined} indicate the best and the second performance. All deep methods are based on VGG19 features.}}
            \label{tab4}
            \centering
            \begin{tabular}{|c|c|c|c|c|c|c|c|c|c|c|c|c|c|c|}
            \hline
                \multicolumn{1}{|c|}{\multirow{2}{*}{Tasks}} & \multicolumn{1}{|c|}{\multirow{2}{*}{Fashions}}& \multicolumn{1}{|c|}{\multirow{2}{*}{Methods}}&
                \multicolumn{4}{c|}{MIRFlickr25K}   & \multicolumn{4}{c|}{NUS-WIDE} & \multicolumn{4}{c|}{Wiki} \\  \cline{4-15}
                && \multicolumn{1}{|c|}{}& 16bits    & 32bits    & 48bits    & 64bits    & 16bits    & 32bits    & 48bits    & 64bits    & 16bits    & 32bits    & 48bits    & 64bits    \\ \hline  \hline
                && SSAH\cite{34}         & 0.780     & 0.791     & 0.793     & 0.795     & 0.621     & 0.630     & 0.623     & 0.615     & 0.253     & 0.306     & 0.317     & 0.327     \\ \cline{3-15}
                && DBRC\cite{35}         & 0.587	    & 0.590     & 0.590     & 0.590	    & 0.394	    & 0.409	    & 0.413     & 0.417 	& 0.253	    & 0.265	    & 0.267     & 0.269     \\ \cline{3-15}
                &Non-continuous& RDCMH\cite{36}        & 0.772 	& 0.774     & 0.777    	& 0.779 	& 0.623 	& 0.624 	& 0.626     & 0.627 	& 0.294 	& 0.297 	& 0.299     & 0.300     \\ \cline{3-15}
                &Hashing& SADCH\cite{37}        & 0.759	    & 0.784     & 0.805	    & 0.826	    & 0.649	    & 0.706	    & 0.730     & 0.753	    & 0.360	    & 0.402	    & 0.406     & 0.410     \\ \cline{3-15}
                && MESDCH\cite{38}       & 0.813 	& 0.830     & 0.834    	& 0.837 	& 0.653 	& 0.670 	& 0.673     & 0.675 	& 0.379 	& 0.371 	& 0.376     & 0.381     \\ \cline{2-15}
                && OCMFH\cite{27}        & 0.635 	& 0.632     & 0.632    	& 0.631 	& 0.412 	& 0.423 	& 0.419     & 0.414 	& 0.177 	& 0.186 	& 0.187     & 0.188     \\ \cline{3-15}
                I$\to$T&Online& OLSH\cite{29}         & 0.671 	& 0.677     & 0.679    	& 0.680 	& 0.595 	& 0.605 	& 0.606     & 0.606 	& 0.243 	& 0.252 	& 0.256     & 0.259     \\ \cline{3-15}
                &Hashing& LEMON\cite{30}        & 0.742     & 0.748     & 0.750 	& 0.752 	& 0.656 	& 0.681 	& 0.676     & 0.671 	& 0.367 	& 0.382 	& 0.399     & 0.416     \\ \cline{3-15}
                && DOCH\cite{32}         & 0.762 	& 0.766     & 0.774    	& 0.781  	& 0.647 	& 0.654 	& 0.658     & 0.662 	& 0.411 	& 0.422 	& 0.423     & 0.424     \\ \cline{2-15}
                && DLCH-1                & \textbf{0.848} & \textbf{0.865} & \textbf{0.872}	& \underline{0.879}	& \underline{0.819}	& 0.831	& 0.835 & 0.838    & \textbf{0.526}  & \textbf{0.537}	& \textbf{0.530} & \textbf{0.523}    \\ \cline{3-15}
                &Lifelong& DLCH-2                & 0.834	    & \underline{0.859}	& \underline{0.870} & \textbf{0.880}	& \textbf{0.821}	& 0.834	& \underline{0.837} & \underline{0.840}    & 0.472	& \underline{0.507}	& \underline{0.494} & \underline{0.480}  \\ \cline{3-15}
                &Hashing& DLCH-3                & 0.834	    & 0.845	    & 0.861     & 0.877	    & 0.808	    & \underline{0.835}	& 0.836 & 0.836     & \underline{0.502}	 & 0.493	& 0.483 & 0.472 \\ \cline{3-15}
                && DLCH-4                & \underline{0.839}	    & 0.857	    & 0.865 & 0.873	    & 0.816	    & \textbf{0.844} & \textbf{0.846}	& \textbf{0.847}    & 0.462 & 	0.465 & 0.470	& 0.475\\ \hline
                && SSAH                  & 0.791	    & 0.800 	& 0.791     & 0.782	    & 0.623	    & 0.627	    & 0.625     & 0.622	    & 0.228 	& 0.283	    & 0.297    & 0.311  \\ \cline{3-15}
                && DBRC                  & 0.588	    & 0.596	    & 0.596     & 0.596	    & 0.425	    & 0.429	    & 0.434     & 0.438  	& 0.544 	& 0.538	    & 0.543    & 0.548 \\ \cline{3-15}
                &Non-continuous& RDCMH                 & 0.749	    & 0.752  	& 0.756     & 0.765 	& 0.605 	& 0.605 	& 0.609     & 0.613 	& 0.293 	& 0.296 	& 0.299    & 0.301     \\ \cline{3-15}
                &Hashing& SADCH                 & 0.773	    & 0.795	    & 0.805     & 0.814	    & 0.680	    & 0.717	    & 0.730     & 0.743	    & 0.624	    & 0.632	    & 0.632    & 0.632     \\ \cline{3-15}
                && MESDCH                & 0.802	    & 0.812 	& 0.815     & 0.818 	& 0.651 	& 0.672 	& 0.676     & 0.679 	& 0.589	    & 0.602 	& 0.605    & 0.608     \\ \cline{2-15}
                && OCMFH                 & 0.711	    & 0.722 	& 0.736     & 0.749 	& 0.423 	& 0.417 	& 0.428     & 0.439 	& 0.410 	& 0.455 	& 0.460    & 0.464     \\ \cline{3-15}
                T$\to$I&Online& OLSH                  & 0.709	    & 0.720 	& 0.729     & 0.738 	& 0.705 	& 0.713 	& 0.717     & 0.720     & 0.507 	& 0.544 	& 0.564    & 0.583     \\ \cline{3-15}
                &Hashing& LEMON                 & 0.816	    & 0.829 	& 0.832     & 0.834 	& 0.778	    & 0.787 	& 0.792     & 0.796 	& 0.641 	& 0.680 	& 0.681    & 0.681     \\ \cline{3-15}
                && DOCH                  & \underline{0.817} 	& \underline{0.832}	& \textbf{0.841}    & \textbf{0.850}	& \textbf{0.779} & \textbf{0.795}	& \textbf{0.803} & \textbf{0.810}	& 0.647	& 0.653	& 0.662 & 0.671    \\ \cline{2-15}
                && DLCH-1                & 0.815	    & 0.827	    & \underline{0.839}     & \textbf{0.850}& 0.761	& 0.782	& 0.790 & 0.797   & \textbf{0.702}	& \textbf{0.729}	  & \textbf{0.728}   & \textbf{0.727}    \\ \cline{3-15}
                &Lifelong& DLCH-2                & \textbf{0.820} & \textbf{0.839} & 0.835   & 0.831	& \underline{0.778}	& \underline{0.790}	& \underline{0.795} & \underline{0.799}   & \underline{0.651}	& 0.689 	& \underline{0.705} & \underline{0.721}\\ \cline{3-15}
                &Hashing& DLCH-3                & 0.814	    & 0.820	    & 0.827     & 0.833	    & 0.753	    & 0.781	    & 0.787     & 0.793     & 0.649	    & 0.690	   & 0.692      & 0.693  \\ \cline{3-15}
                && DLCH-4                & 0.813	    & 0.820 	& 0.821     & 0.822	    & 0.767	    & 0.787	    & 0.790     & 0.792     & 0.632 	& \underline{0.698}	 & 0.694 & 0.690  \\ \hline
            \end{tabular}
            \end{table*}

            \begin{figure*}[!t]
                \centering
                \includegraphics[width=1\linewidth]{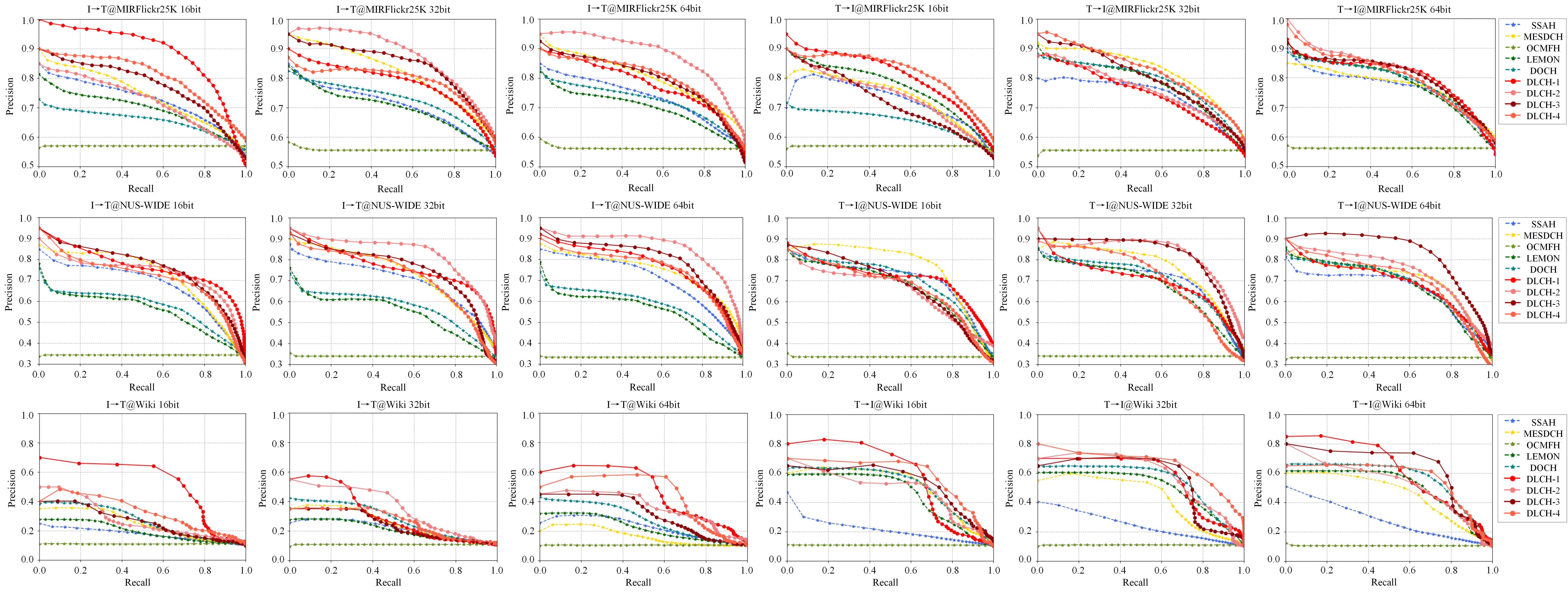}
                \caption{\small{Precision-Recall Curves Evaluated on MIRFilckr, NUS-WIDE and Wiki datasts.}}
                \label{Fig2}
            \end{figure*}

    \subsection{Implementation Details and Evaluations}
        All experiments were conducted on Intel(R) Xeon(R) Gold6148CPU with 20 cores and eight Tesla V100-SXM2 GPUs with using the parameter setting in original papers to ensure impartiality and objectivity. The batchsize is set to 64, and the epoch is fixed to 2,000. Inspired by [39], [40], initial learning rate in original and incremental hashing learning stage are 0.001 and 0.000001, respectively. All the hyper-parameters will be set automatically in the range of $[10^{-2}, 10^5]$ according cross-validation results. A detailed analysis and comparison of parameter sensitivity will be provided.

        In order to objectively evaluate the performance of DLCH, we adopt two widely-used evaluations, including Mean Average Precision (MAP) and Precision-Recall Curves Hamming distance 2. Then, we compare our method with nine state-of-the-art hashing methods, including SSAH\cite{34}, DBRC\cite{35}, RDCMH\cite{36}, SADCH\cite{37}, MESDCH\cite{38}, OCMFH\cite{27}, OLSH\cite{29}, LEMON\cite{30}and DOCH\cite{32}. The first five methods are non-continuous deep cross-modal hashing methods and the rest are online cross-modal hashing methods. Some have kindly provided the source codes, we refer to parameters settings in original papers.

    \subsection{Results of hash retrieval}
        The MAP comparisons of DLCH with different hash bits on three benchmark datasets are reported in Table V, from which we can see that DLCH achieves better performance compared to other baselines in most cases. In Table III, DLCH-$i$ denotes that the number of incremental classes.

        Comparing with the optimal non-continuous baseline (i.e. MESDCH and SADCH) when using image to retrieve text (i.e., I$\to$T), ours achieves average increments of 3.1\%, 12.8\% and 10.4\% on MIRFlickr25K, NUS-WIDE and Wiki, respectively. When using text to retrieve image (i.e., T$\to$I), ours also achieves average increments of 4.5\%, 6.8\%, and 6.7\% on MIRFlickr25K, NUS-WIDE and Wiki, respectively. Then, comparing with the online methods, ours also boosts the performance significantly, especially on I$\to$T task. To be specific, ours reaches average increments of 8.8\%, 16.2\% comparing DOCH which performs the best on MIRFlickr25K and NUS-WIDE. Additionally, when evaluating on Wiki, ours achieve significantly gains average increments of 21.6\% and 20.2\% on I$\to$T and T$\to$I task, respectively.

        Apart from MAP evaluations, we also illustrate precision-recall curves with different hash code lengths of comparative methods on benchmark datasets in Fig.2, from which it can be observed that ours achieve the best score on each length, which is consistent with observations on the MAP scores. Note that DLCH does not achieve best score on each dataset, and it incurs minor drops on NUS-WIDE comparing with DOCH when using text to retrieve image. It can be analyzed that no incremental data with new category appears, which makes comparisons in Table III and Fig.2 better than they would have been if the new categories had appeared. Combing all the results in Table III and Fig.2, we can conclude that our proposed method outperforms than most of recent state-of-the-art cross-modal hashing methods, including non-continuous and online hashing. Then, the larger is the incremental classes, the higher is the performance.

    \subsection{Catastrophic Forgetting Analysis}
        As mentioned before, almost all the existing cross-modal hashing method is suffering from catastrophic forgetting when when adding the incremental data with new categories and using trained hash functions to retrieve the incremental data. We now provide some results on catastrophic forgetting evaluations as shown in Fig.3.

        From Fig.3, it can be clearly seen that all the non-continuous cross-modal hashing methods suffer from serve performance degradation due to the catastrophic forgetting. Moreover, when there are more new categories, the performance degrades seriously. Then, we have to note that the online learning process alleviates the phenomenon to some extent. However, it still incurs unacceptable drops. Taking DOCH as an example shown in Fig.5, for each additional category of data, the performance will respectively decrease about by 3.2\% and 3.7\% on I$\to$T and T$\to$I task, respectively. Comparing with these methods, our DLCH basically maintains the performance and sometimes even improves as the number of new categories increases, which benefits from the proposed lifelong hashing learning strategy.

        \begin{figure*}[!t]
                \centering
                \includegraphics[width=1\linewidth]{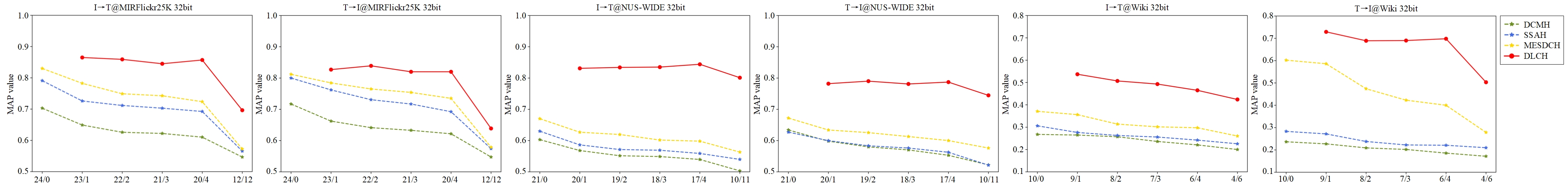}
                \caption{\small{Results on catastrophic forgetting evaluations.}}
                \label{Fig3}
        \end{figure*}

   \subsection{Time cost analysis}
        Apart from retrieval accuracy and catastrophic forgetting, training time and resource cost are also concerned. As analyzed before, the complexity of incremental hash codes learning in our method is linearly dependent on the size of incremental data ${n}$. Thus, we further conduct experiments to demonstrate it, and the results are shown in Table IV.
            \begin{table}[!t]
            \caption{\small{Training Time Comparisons (in Minutes) of Baselines on Benchmark Datasets with One Incremental Class and 32 bits.}}
            \label{tab6}
            \centering
            \begin{tabular}{|l|l|l|l|l|}
            \hline
            \multicolumn{1}{|c|}{Learning Fashion}                                            & \multicolumn{1}{c|}{Methods}      & MIRFlicker & NUS-WIDE & WIKI \\ \hline
            \multirow{6}{*}{\begin{tabular}[c]{@{}l@{}}Non-continuous\\ Hashing\end{tabular}} & DCMH\cite{39}                              & 227.8      & 17170.1  & 20.6 \\ \cline{2-5}
                                                                                              & SSAH                              & 431.2      & 22000.4  & 25.8 \\ \cline{2-5}
                                                                                              & DBRC                              & 358.8      & 20681.0  & 35.4 \\ \cline{2-5}
                                                                                              & AADCMH\cite{40}                             & 447.8      & 30488.6  & 45.2 \\ \cline{2-5}
                                                                                              & AGCN\cite{41}                               & 289.3      & 21085.9  & 34.6 \\ \cline{2-5}
                                                                                              & MESDCH                           & 337.8      & 20201.1  & 25.4 \\ \hline
            \multirow{5}{*}{Online Hashing}                                                   & OCMFH                             & 32.6       & 310.1    & 2.3  \\ \cline{2-5}
                                                                                              & OLSH                              & 33.9       & 297.6    & 2.1  \\ \cline{2-5}
                                                                                              & LEMON                             & 46.3       & 360.1    & 2.3  \\ \cline{2-5}
                                                                                              & DOCH                              & 41.2       & 357.2    & 2.5  \\ \cline{2-5}
                                                                                              & OMGH                              & 41.6       & 393.1    & 2.6  \\ \hline
            Lifelong Hashing                                                                  & DLCH                              & 76.1       & 1016.2   & 6.7  \\ \hline
            \end{tabular}
            \end{table}

        As shown in Table IV, our DLCH possesses absolute advantage comparing with non-continuous methods, and the advantage is more obvious on larger scale dataset. When incremental data is added to the database, these baseline methods need to use all the accumulated data for retraining, which requires more training time in repeat way. On the contrary, DLCH only utilizes incremental data to update hash functions, which significantly reduces training time and resource cost. It can be also observed that ours need more training time comparing with other online methods, which results from that our DLCH is deep learning-based methods. Combining with the above experimental results, we can find that DLCH has excellent time performance, which is more significant on large-scale datasets. Considering the high retrieval accuracy and low performance performance degradation, the proposed DLCH possesses strong competitiveness comparing with these cross-modal hashing methods.

    \subsection{Sensitivity to parameters}
        There are five hyper-parameters in DLCH, where $\alpha$, $\beta$ and $\gamma$ are designed for original learning phase and $\lambda$ and $\mu$ are used in lifelong learning phase. We now  evaluate the influence of hyper-parameters in controlling weight ratio among losses. We firstly set the hash code length $k = 64$. Then, we calculate MAP values by adjusting the parameters between 10$^{-2}$ and 10$^{5}$ with a multiplication step of 10. Fig.4 shows experimental results of hyper-parameters $\alpha$, $\beta$, $\gamma$, $\lambda$ and $\mu$ on MIRFlickr.

            \begin{figure*}[!t]
                \centering
                \includegraphics[width=1\linewidth]{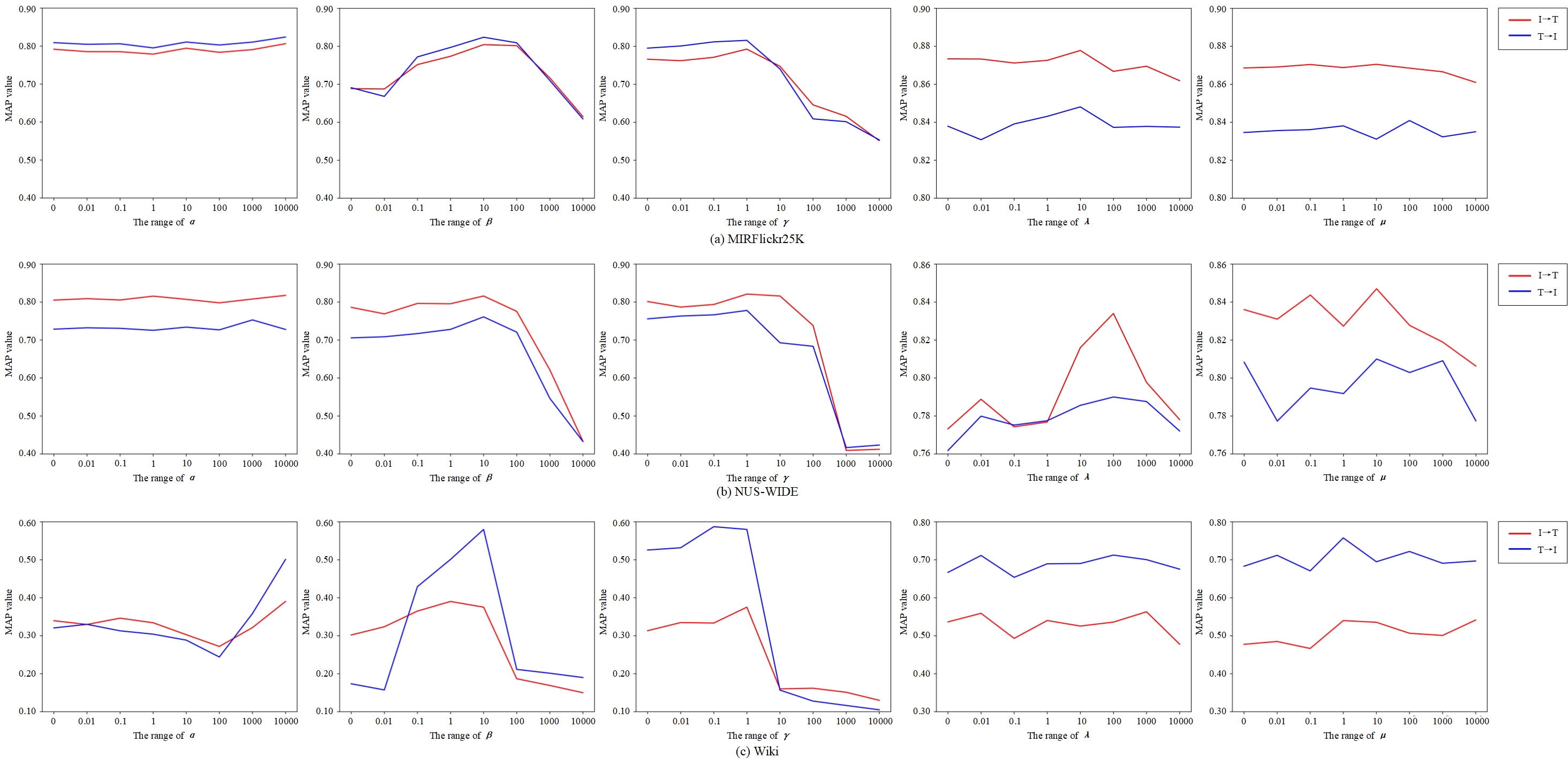}
                \caption{\small{Sensitivity analysis on MIRFlickr Dataset in the range of $[10^{-2}, 10^5]$.}}
                \label{Fig5}
            \end{figure*}

        As illustrated in Fig.4, different $\beta$ and $\gamma$ have more obvious impact on the MAP value, while it changes more gently with $\alpha$. During incremental hashing learning, $\lambda$ and $\mu$ do not have particularly great impact on MAP values. According to the similar way, the best optimum values for three benchmark datasets are summarized in TABLE V.
            \begin{table}[!t]
            \caption{\small{Optimum Parameter of Different Tasks on Benchmark Datasets.}}
            \centering
            \begin{tabular}{|c|l|l|l|l|l|l|}
            \hline
            \multicolumn{1}{|c|}{\multirow{2}{*}{Parameters}} & \multicolumn{2}{c|}{MIRFlicker}    & \multicolumn{2}{c|}{NUS-WIDE}   & \multicolumn{2}{c|}{WIKI}     \\ \cline{2-7}
            \multicolumn{1}{|c|}{}                            & T$\to$I          & I$\to$T         & T$\to$I          & I$\to$T      & T$\to$I           & I$\to$T   \\ \hline\hline
                $\alpha$                                      & 10               & 10              & 10000            & 10000        & 10000             & 10000     \\ \hline
                $\beta$                                       & 10               & 100             & 12               & 10           & 7                 & 9         \\ \hline
                $\gamma$                                      & 1                & 1               & 1                & 1            & 0.6               & 0.8       \\ \hline
                $\lambda$                                     & 10               & 10              & 200              & 200          & 1000              & 200       \\ \hline
                $\mu$                                         & 10               & 100             & 50               & 50           & 1                 & 1         \\ \hline
            \end{tabular}
            \end{table}

    \subsection{Ablation experiment}
        In order to achieve lifelong hashing retrieval, several losses are defined. We now investigate the variants of DLCH to further analyze effectiveness of each loss. During the exploration, we compare these variants on MIRFlickr dataset with setting the number of incremental class be 1.

        Firstly, we design three variants, DLCH-intra, DLCH-inter and DLCH-quant, to make sure original hashing loss function. DLCH-intra, DLCH-inter and DLCH-quant are the DLCH variant which remove intra-modality, inter-modality similarity preserving and quantization loss, respectively. Then, to explore the effectiveness of lifelong hashing, four additional variants are designed. DLCH-O, DLCH-L, DLCH-Q and DLCH-B are the variants without using original similarity preserving loss, lifelong hashing loss, quantization loss and bit balance loss, respectively. The comparisons are shown Table VI.
            \begin{table}[!t]
                \caption{\small{MAP Comparison in Original Hashing.}}
                \label{tab6}
                \centering
                \begin{tabular}{|c|c|c|c|c|c|c|}
                \hline
                    \multicolumn{1}{|c|}{\multirow{2}{*}{Methods}} & \multicolumn{3}{c|}{$I \to T$}    & \multicolumn{3}{c|}{$T \to I$}   \\  \cline{2-7}
                    \multicolumn{1}{|c|}{}            & 16bits    & 32bits   & 64bits & 16bits & 32bits & 64bits  \\
                    \hline\hline
                        DLCH                              & 0.848     & 0.865    & 0.879  & 0.815  & 0.827  & 0.850 \\
                    \hline\hline
                        DLCH-intra                        & 0.823     & 0.841    & 0.837  & 0.743  & 0.717  & 0.735 \\
                    \hline
                        DLCH-inter                        & 0.808     & 0.828    & 0.836  & 0.739  & 0.730  & 0.732 \\
                    \hline
                        DLCH-quant                        & 0.810     & 0.835    & 0.860  & 0.791  & 0.819  & 0.839 \\
                    \hline\hline
                        DLCH-O                            & 0.841     & 0.851    & 0.872  & 0.811  & 0.822  & 0.838 \\
                    \hline
                        DLCH-I                            & 0.836     & 0.855    & 0.877  & 0.806  & 0.819  & 0.836 \\
                    \hline
                        DLCH-Q                            & 0.833     & 0.852    & 0.877  & 0.803  & 0.810  & 0.837 \\
                    \hline
                        DLCH-B                            & 0.835     & 0.851    & 0.879  & 0.811  & 0.820  & 0.832 \\
                    \hline
                \end{tabular}
            \end{table}

        During the original hashing learning, it can be seen from Table VI that DLCH achieves average increments of 4.1\% and 3.7\% on all bits over DLCH-intra which does not consider the similarity among intra modalities these two tasks. DLCH gains average increments of 7.3\% and 8.1\% over DLCH-inter which ignores inter-similarity. Then, DLCH-quant uses continuous relaxation to address discrete optimization and results, which naturally incurs some drops comparing with DLCH. During the incremental hashing learning, due to the introduction of intra-modality and inter-similarity similarity preserving, and quantization constraints, DLCH also gains clear advantages over DLCH-O, DLCH-L, DLCH-Q and DLCH-B. Lifelong hashing loss is designed to keep the trained hash function continuously usable by preserving the similarity between original and incremental data. From Table VI, we can see that DLCH achieves higher increments, i.e., 6.7\% and 7.5\% comparing with DLCH-O, which means that it indeed preserves performance and avoids catastrophic forgetting.

        Furthermore, in order to verify the validity of multi-label semantic similarity, we compare the variants which use multi- and single-label semantic similarity, and the comparisons are shown in Table VII. From Table VII, it can be seen that DLCH with multi-label semantic similarity improves retrieval accuracy by $20\%$ comparing with single-label semantic similarity, which is consistent with the analysis in Section III-D.
            \begin{table}[!t]
                \caption{\small{MAP Comparison of Label Semantic Similarity.}}
                \label{tab6}
                \centering
                \begin{tabular}{|c|c|c|c|c|c|c|}
                \hline
                    \multicolumn{1}{|c|}{\multirow{2}{*}{Methods}} & \multicolumn{3}{c|}{$I \to T$}                    & \multicolumn{3}{c|}{$T \to I$}   \\  \cline{2-7}
                    \multicolumn{1}{|c|}{}            & 16bits    & 32bits   & 64bits    & 16bits  & 32bits  & 64bits  \\
                    \hline\hline
                    DLCH-multi                        & 0.848     & 0.865    & 0.879     & 0.815   & 0.827   & 0.850 \\
                    \hline
                    DLCH-single                       & 0.823     & 0.841    & 0.837     & 0.743   & 0.717   & 0.735 \\
                    \hline
                \end{tabular}
            \end{table}

\section{Conclusion}
    In this paper, we propose a novel deep lifelong cross-modal hashing, which effectively solves the problem brought by new categories appearing in database. Our method is divided into two phases: original learning and lifelong learning. We employ a multi-label semantic similarity for the first phase to supervise the learning of hash functions and obtain high-quality original hash codes. Meanwhile, we design a lifelong hashing loss for the second phase, where incremental hash codes are directly learned while original hash code remains unchanged. Extensive experiments on three well-known real-world benchmarks show that the proposed method has better performance than other baseline methods with reducing training time significantly and is an effective cross-modal hashing retrieval method.

\section*{Acknowledgment}
    The authors would like to thank the anonymous reviewers for their help. This work was supported by the National Natural Science Foundation of China (Grant No. 62176217, 62206224), the Natural Science Foundation of Sichuan Province (Grant No. 2022NSFSC0866), the Innovation Team Funds of China West Normal University (Grant No. KCXTD2022-3), and the Doctoral Research Innovation Project (Grant No. 21E025).


\begin{thebibliography}{60}
    \bibitem{1}
    P. Kaur, H. Pannu and A. Malhi, ``Comparative analysis on cross-modal information retrieval: A review,'' in \emph{Comput. Sci. Rev.}, vol. 39, pp. 100336, 2021.
\bibitem{2}
    W. Cao, W. Feng, Q. Lin, G. Cao and Z. He, ``A Review of Hashing Methods for Multimodal Retrieval,'' in \emph{IEEE Access}, vol. 8, pp. 15377-15391, 2020.
\bibitem{3}
    G. Menghani, ``Efficient Deep Learning: A Survey on Making Deep Learning Models Smaller, Faster, and Better,'' in \emph{ACM Comput. Surv.}, vol. 55, no. 12, pp. 1-37, 2023.
\bibitem{4}
    M. Lange, R. Aljundi, M. Masana, S. Parisot, X. Jia, A. Leonardis, G. Slabaugh and T. Tuytelaars, ``A Continual Learning Survey: Defying Forgetting in Classification Tasks,'' in \emph{IEEE Trans. Pattern Anal. Mach. Intell.}, vol. 44, pp. 3366-3385, 2021.
\bibitem{5}
    X. Luo, C. Chen, H. Zhong, H. Zhang, M. Deng, J. Huang and X. Hua, ``A Survey on Deep Hashing Methods,'' \emph{ACM Trans. Knowl. Discov. Data.}, vol. 17, no. 1, pp. 1-50, 2023.
\bibitem{6}
    Y. Peng, H. Xin, and Y. Zhao, ``An Overview of Cross-Media Retrieval: Concepts, Methodologies, Benchmarks and Challenges,'' \emph{IEEE Trans. Cir. Syst. Video Techn.}, vol. 28, no. 9, pp. 2372-2385, 2019.
\bibitem{7}
    T. Hoang, T. Do, T. Nguyen and N. Cheung, ``Unsupervised Deep Cross-modality Spectral Hashing,'' \emph{IEEE Trans. Image Process.}, vol. 29, pp. 8391-8406, 2020.
\bibitem{8}
    S. Liu, S. Qian, Y. Guan, J. Zhan and L. Ying, ``Joint-modal Distribution-based Similarity Hashing for Large-scale Unsupervised Deep Cross-modal Retrieval,'' in \emph{Proc. Int. ACM SIGIR Conf. Res. Dev. Inf. Retr.}, 2020, pp. 1618-1625.
\bibitem{9}
    J. Yu, H. Zhou, Y. Zhan and D. Tao, ``Deep Graph-neighbor Coherence Preserving Network for Unsupervised Cross-modal Hashing,'' \emph{Proc. AAAI Conf. Artif. Intell.}, 2021, pp. 4626-4634.
\bibitem{10}
    Y. Wang, B. Xue, Q. Cheng, Y. Chen and L. Zhang, ``Deep Unified Cross-Modality Hashing by Pairwise Data Alignment,'' \emph{Proc. Int. Joint Conf. Artif. Intell.}, 2021, pp. 1129-1135.
\bibitem{11}
    M. Li and H. Wang, ``Unsupervised Deep Cross-Modal Hashing by Knowledge Distillation for Large-scale Cross-modal Retrieval,'' in \emph{Proc. Int. Conf. Multimedia Retr.}, 2021, pp. 183-191.
\bibitem{12}
    T. Hoang, T. Do, T. Nguyen and N. Cheung, ``Multimodal Mutual Information Maximization: A Novel Approach for Unsupervised Deep Cross-Modal Hashing,'' in \emph{IEEE Trans. Neural Networks Learn. Syst.}, pp. 1-14, 2022.
\bibitem{13}
    R. Tu, J. Jiang, Q. Lin, C. Cai, S. Tian, H. Wang and W. Liu, ``Unsupervised Cross-modal Hashing with Modality-interaction,'' in \emph{IEEE Trans. Circuits Syst. Video Technol.}, 2023.
\bibitem{14}
    H. Yao, Y. Zhan, Z. Chen, X. Luo and X. Xu, ``TEACH: Attention-Aware Deep Cross-Modal Hashing,'' in \emph{Proc. Int. Conf. Multimedia Retr.}, 2021, pp. 376-384.
\bibitem{15}
    X. Zou, S. Wu, N. Zhang and E. Bakker, ``Multi-label modality enhanced attention based self-supervised deep cross-modal hashing,'' in \emph{Knowl. Based Syst.}, vol. 239, pp. 107927, 2022.
\bibitem{16}
    F. Yang, Y. Liu, X. Ding, F. Ma and J. Cao, ``Asymmetric cross¨Cmodal hashing with high¨Clevel semantic similarity,'' in \emph{Pattern Recognit.}, vol. 130, pp. 108823, 2022.
\bibitem{17}
    X. Liu, H. Zeng, Y. Shi, J. Zhu, C. Hsia and K. Ma, ``Deep Cross-modal Hashing Based on Semantic Consistent Ranking,'' in \emph{IEEE Trans. Multimedia}, pp. 1-12, 2023.
\bibitem{18}
    Z. Wang, Q. She and T. Ward, ``Generative Adversarial Networks in Computer Vision: A Survey and Taxonomy,'' in \emph{ACM Comput. Surv.}, vol. 54, no. 2, pp. 1-38, 2021.
\bibitem{19}
    X. Ma, T. Zhang, and C. Xu, ``Multi-Level Correlation Adversarial Hashing for Cross-Modal Retrieval,'' \emph{IEEE Trans. Multimedia}, vol. 20, no. 4, pp. 1224-1237, 2020.
\bibitem{20}
    Xi Zhang and Hanjiang Lai and Jiashi Feng, ``Deep Adversarial Discrete Hashing for Cross-Modal Retrieval,'' in \emph{Proc. ACM SIGMM Int. Conf. Multimedia Retr.}, 2020, pp. 525-531.
\bibitem{21}
    J. Zhang and Y. Peng, ``Multi-Pathway Generative Adversarial Hashing for Unsupervised Cross-Modal Retrieval,'' in \emph{IEEE Trans. Multimedia}, vol. 22, pp. 174-187, 2020.
\bibitem{22}
    D. Xie, C. Deng, C. Li, X. Liu and D. Tao, ``Multi-Task Consistency-Preserving Adversarial Hashing for Cross-Modal Retrieval,'' in \emph{IEEE Trans. Image Process.}, vol. 29, pp. 3626-3637, 2020.
\bibitem{23}
    S. Qian, D. Xue, H. Zhang, Q. Fang and C. Xu, ``Dual Adversarial Graph Neural Networks for Multi-label Cross-modal Retrieval,'' in \emph{Proc. AAAI Conf. Artif. Intell.}, 2021, pp. 2440-2448.
\bibitem{24}
    M. Li, Q. Li, Y. Ma and D. Yang, ``Semantic-guided autoencoder adversarial hashing for large-scale cross-modal retrieval,'' in \emph{Complex Intell. Syst.}, vol. 8, pp. 1603-1617, 2022.
\bibitem{25}
    J. Li, E. Yu, J. Ma, X. Chang, H. Zhang and J. Sun, ``Discrete Fusion Adversarial Hashing for cross-modal retrieval,'' in \emph{Knowledge-Based Syst.}, vol. 253, pp. 109503, 2022.
\bibitem{26}
    W. Ou, J. Deng, L. Zhang, J. Gou and Q. Zhou, ``Cross-Modal Generation and Pair Correlation Alignment Hashing,'' in \emph{IEEE Trans. Intell. Transp. Syst.}, vol. 24, pp. 3018-3026, 2023.
\bibitem{27}
    D. Wang and Q. Wang, Y. An, X. Gao and Y. Tian, ``Online Collective Matrix Factorization Hashing for Large-Scale Cross-Media Retrieval,'' in \emph{Proc. Int. ACM SIGIR Conf. Res. Deve. Inf. Retr.}, 2020, pp. 1409-1418.
\bibitem{28}
    L. Xie, J. Shen and L. Zhu, ``Online Cross-Modal Hashing for Web Image Retrieval,'' in \emph{Proc. AAAI Conf. Artif. Intell.}, 2016, pp. 294-300.
\bibitem{29}
    T. Yao, G. Wang, L. Yan, X. Kong, Q. Su, C. Zhang and Q. Tian, ``Online latent semantic hashing for cross-media retrieval,'' in \emph{Pattern Recognit.}, vol. 89, pp. 1-11, 2019.
\bibitem{30}
    Y. Wang, X. Luo and X. Xu, ``Label Embedding Online Hashing for Cross-Modal Retrieval,'' \emph{Proc. ACM Int. Conf. Multimedia}, 2020, pp. 871-879.
\bibitem{31}
    J. Yi, X. Liu, Y. Cheung, X. Xu, W. Fan and Y. He, ``Efficient Online Label Consistent Hashing for Large-Scale Cross-Modal Retrieval,'' in \emph{Proc. IEEE Int. Conf. Multimedia Expo}, 2021, pp. 1-6.
\bibitem{32}
    Y. Zhan, Y. Wang, Y. Sun, X. Wu, X. Luo and X. Xu, ``Discrete online cross-modal hashing,'' \emph{Pattern Recognit.}, vol. 122, pp. 108262, 2022.
\bibitem{33}
    X. Liu, J. Yi, Y. Cheung, X. Xu and Z. Cui, ``OMGH: Online Manifold-Guided Hashing for Flexible Cross-modal Retrieval,'' in \emph{IEEE Trans. Multimedia}, 2022.
\bibitem{34}
    C.Li, C. Deng, N. Li, W. Liu, X. Gao, and D. Tao, ``Self-supervised adversarial hashing networks for cross-modal retrieval, '' \emph{Proc. IEEE Conf. Comput. Vis. Pattern Recognit.}, 2018, pp. 4242-4251.
\bibitem{35}
    D. Hu, F. Nie, and X. Li, ``Deep Binary Reconstruction for Cross-Modal Hashing,'' in \emph{IEEE Trans. Multimedia}, vol. 21, no. 4, pp. 973-985, 2019.
\bibitem{36}
    X. Liu, G. Yu, C. Domeniconi, J. Wang, Y. Ren and M. Guo, ``Ranking-based Deep Cross-modal Hashing,'' in \emph{Proc. AAAI Conf. Artif. Intell.}, 2019, pp. 4400-4407.
\bibitem{37}
    Y. Wang, X. Shen, Z. Tang, T. Zhang and J. Lv, "Semi-Paired Asymmetric Deep Cross-Modal Hashing Learning," in \emph{IEEE Access}, vol. 8, pp. 113814-113825, 2020.
\bibitem{38}
    X. Zou, S. Wu, M. B. Erwin and X. Wang, ``Multi-label enhancement based self-supervised deep cross-modal hashing,'' in \emph{Neurocomputing}, vol. 467, pp. 138-162, 2022.
\bibitem{39}
    Q. Jiang and W. Li, ``Deep Cross-Modal Hashing,'' in \emph{Proc. IEEE Conf. Comput. Vis. Pattern Recognit.}, 2017, pp. 3232-3240.
\bibitem{40}
\bibitem{41}
    X. Dong, L. Liu, L. Zhu, L. Nie and H. Zhang, "Adversarial Graph Convolutional Network for Cross-Modal Retrieval," in \emph{IEEE Trans. Circuits Syst. Video Technol.}, vol. 32, pp. 1634-1645, 2021.
\end{thebibliography}
\end{document}